\documentclass{article}

\PassOptionsToPackage{authoryear,round,sort,comma}{natbib}
 \usepackage[preprint]{neurips_2026}

\usepackage[utf8]{inputenc} 
\usepackage[T1]{fontenc}    
\usepackage[colorlinks=true,citecolor=blue,breaklinks,backref=page]{hyperref} 
\usepackage{url}            

\renewcommand*{\backref}[1]{}
\renewcommand*{\backrefalt}[4]{%
  \ifcase #1 (Not cited)%
  \or (Cited on page~#2)%
  \else (Cited on pages~#2)%
  \fi%
}

\hypersetup{
    colorlinks,
    citecolor=[rgb]{0.00, 0.45, 0.70}, %
    linkcolor=[rgb]{0.7, 0, 0.4},
    urlcolor=[rgb]{0.7, 0, 0.4} %
}

\usepackage{booktabs}       
\usepackage{amsfonts}       
\usepackage{nicefrac}       
\usepackage{microtype}      
\usepackage{xcolor}         

\usepackage{graphicx}
\usepackage{subcaption}      
\usepackage{array}
\usepackage{arydshln}

\usepackage{amsmath}
\usepackage{amssymb}
\usepackage{mathtools}
\usepackage{amsthm}
\usepackage{multirow}
\usepackage[capitalize,noabbrev]{cleveref}
\usepackage[textsize=tiny]{todonotes}
\usepackage{tikz}
\usepackage{xspace}

\usepackage{todonotes}
\usepackage[normalem]{ulem}
\usepackage{xspace}

\makeatletter
\newcommand*\iftodonotes{\if@todonotes@disabled\expandafter\@secondoftwo\else\expandafter\@firstoftwo\fi}  
\makeatother

\newcommand{\model}{\mathcal{M}}
\newcommand{\task}{\mathcal{T}}

\newcommand{\traj}{\mathbf{y}}
\newcommand{\promptx}{\mathbf{x}}

\newcommand{\limitbanner}[3]{
  \AtBeginShipout{\ifnum\value{page}=#1%
    \AtBeginShipoutUpperLeft{\put(0,\dimexpr-\paperheight+1.5em\relax){%
      \colorbox{#2}{\makebox[\dimexpr\paperwidth-2\fboxsep\relax][c]{%
        \color{white}\sffamily\bfseries\small #3}}}}%
  \fi}%
}

\newcommand*{\circled}[1]{\tikz[baseline=(char.base)]{
          \node[shape=circle,draw,line width=1.0pt,inner sep=1pt] (char) {\normalfont{\small #1}};}}

\title{Forecasting Downstream Performance of LLMs\\ With Proxy Metrics}

\newcommand{\github}{\raisebox{-1.5pt}{\includegraphics[height=0.9em]{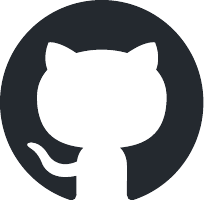}}\xspace}

\newcommand{\mila}{\raisebox{1.0ex}{\includegraphics[height=1.8ex, trim=10 0 0 0, clip]{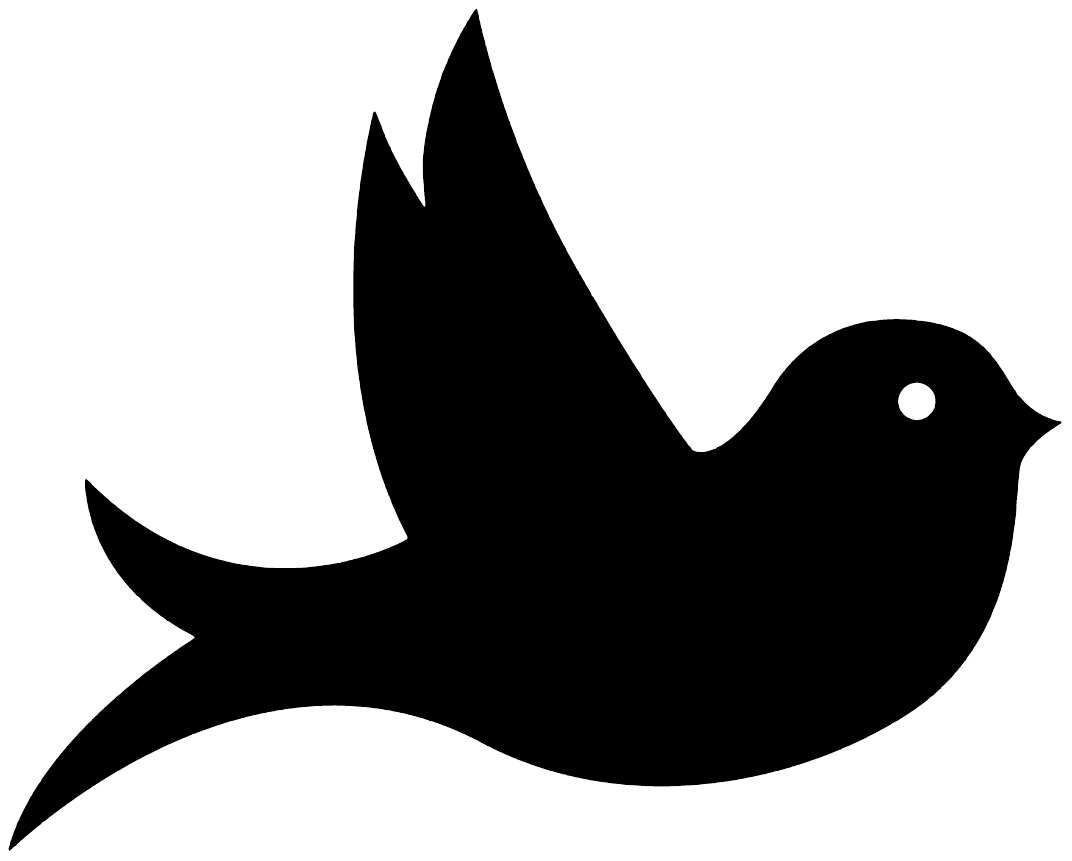}}\,}

\newcommand{\snow}{\raisebox{1.05ex}{\small$\Omega$}\,}

\newcommand{\periodic}{\raisebox{1.8ex}{\tikz[baseline, scale=0.015]{
  \fill[black] (0,0) circle (3.0);
  \draw[line width=0.5pt, black] (0,0) circle (8);
  \fill[black] (7.2,3.4) circle (2.4);
  \fill[black] (-5.7,-5.7) circle (2.4);
}}\,}

\newcommand{\cifar}{\raisebox{2.6ex}{\tikz[baseline, x=0.00030cm, y=-0.00030cm]{
\path[fill=black,draw=none]
(480.00,0.48) --
(558.62,155.00) -- (560.68,158.00) -- (563.00,160.22) --
(567.00,162.43) -- (570.00,163.25) -- (574.00,163.36) --
(578.00,162.38) -- (652.00,119.71) -- (657.00,116.88) --
(657.92,117.00) -- (609.64,366.00) -- (609.39,370.00) --
(609.83,374.00) -- (611.64,378.00) -- (613.00,379.68) --
(617.00,383.16) -- (620.00,384.48) -- (624.00,385.25) --
(627.00,385.08) -- (630.00,384.37) -- (635.00,381.36) --
(726.69,283.00) -- (736.00,273.15) -- (737.00,272.84) --
(762.72,333.00) -- (767.00,337.52) -- (770.00,339.28) --
(773.00,340.20) -- (779.00,340.25) -- (907.00,313.04) --
(907.80,314.00) -- (907.31,316.00) -- (864.58,447.00) --
(863.66,450.00) -- (863.36,455.00) -- (863.74,458.00) --
(865.54,462.00) -- (868.12,465.00) -- (871.00,467.21) --
(921.00,490.61) -- (921.70,491.00) -- (921.80,492.00) --
(701.00,670.79) -- (695.56,676.00) -- (693.84,679.00) --
(692.75,683.00) -- (692.67,686.00) -- (693.53,690.00) --
(721.00,766.09) -- (720.00,766.35) -- (714.00,765.32) --
(517.00,730.58) -- (509.00,730.54) -- (505.00,731.60) --
(501.00,733.57) -- (498.00,735.72) -- (495.71,738.00) --
(493.56,741.00) -- (491.59,745.00) -- (490.52,749.00) --
(490.31,754.00) -- (500.85,959.00) -- (458.15,959.00) --
(468.69,754.00) -- (468.47,749.00) -- (467.42,745.00) --
(465.44,741.00) -- (463.27,738.00) -- (460.00,734.94) --
(454.00,731.61) -- (450.00,730.54) -- (442.00,730.58) --
(246.00,765.21) -- (239.00,766.40) -- (238.00,766.12) --
(265.47,690.00) -- (266.33,686.00) -- (266.24,683.00) --
(265.14,679.00) -- (263.45,676.00) -- (257.00,669.94) --
(37.24,492.00) -- (37.30,491.00) -- (38.00,490.61) --
(88.00,467.28) -- (91.00,465.00) -- (93.45,462.00) --
(95.24,458.00) -- (95.63,455.00) -- (95.49,451.00) --
(94.43,447.00) -- (51.69,316.00) -- (51.19,314.00) --
(52.00,313.04) -- (180.00,340.25) -- (186.00,340.20) --
(191.00,338.28) -- (194.00,335.74) -- (196.27,333.00) --
(222.00,272.83) -- (223.00,273.10) -- (233.27,284.00) --
(319.76,377.00) -- (325.00,382.23) -- (329.00,384.37) --
(333.00,385.21) -- (336.00,385.16) -- (339.00,384.47) --
(342.00,383.15) -- (346.00,379.67) -- (347.35,378.00) --
(349.15,374.00) -- (349.60,370.00) -- (349.35,366.00) --
(301.00,117.00) -- (302.00,116.79) -- (381.00,162.38) --
(385.00,163.36) -- (389.00,163.24) -- (392.00,162.42) --
(396.00,160.21) -- (398.35,158.00) -- (400.38,155.00) --
(478.66,1.00) -- (479.00,0.49) -- cycle;
}}\,}

\author{
\vspace{2mm}
        Arkil Patel~\mila\hspace{4.5mm}
        Siva Reddy~\mila\cifar\snow \hspace{4.5mm}
        Marius Mosbach~\mila \hspace{4.5mm}
        Dzmitry Bahdanau~\mila\cifar\periodic
    \vspace{3mm}
    \\
    \mila Mila -- Quebec AI Institute \& McGill University
    \vspace{2mm}
    \\
    \cifar Canada CIFAR AI Chair
    \hspace{4mm}  
    \snow ServiceNow Research
    \hspace{4mm}  
    \periodic Periodic Labs
    \vspace{4mm}
    \\
    Correpondence to: \texttt{\href{mailto:arkil.patel@gmail.com}{arkil.patel@gmail.com}} \hspace{5mm} \github \href{https://github.com/McGill-NLP/proxy-metrics}{McGill-NLP/proxy-metrics}
}

\begin{document}

\maketitle

\begin{abstract}

Progress in language model development is often driven by comparative decisions: which architecture to adopt, which pretraining corpus to use, or which training recipe to apply. 
Making these decisions well requires reliable performance forecasts, yet the two commonly used signals are fundamentally limited. 
Cross-entropy loss is poorly aligned with downstream capabilities, and direct downstream evaluation is expensive, sparse, and often uninformative at early training stages. 
Instead, we propose to construct proxy metrics by aggregating token-level statistics, such as entropy, top-k accuracy, and expert token rank, from a candidate model's next token distribution over expert-written solutions.
Across three settings, our proxies consistently outperform loss- and compute-based baselines: 1) For cross-family model selection, they rank a heterogeneous population of reasoning models with mean Spearman $\rho = 0.81$ (vs. $\rho = 0.36$ for cross-entropy loss); 2) For pretraining data selection, they reliably rank 25 candidate corpora for a target model at roughly $10{,}000\times$ less compute than direct evaluation, pushing the Pareto frontier beyond existing methods; and 3) for training-time forecasting, they extrapolate downstream accuracy across an $18\times$ compute horizon with roughly half the error of existing alternatives. 
Together, these results suggest that expert trajectories are a broadly useful source of signal for assessing model capabilities, enabling reliable performance forecasting throughout the model development life cycle.

\end{abstract}

\section{Introduction}

Large language model (LLM) development requires making \emph{comparative} decisions: which pretraining corpus is better, which post-training recipe increases performance on a target domain, and whether a new model architecture is better than the current frontier. 
A common signal for resolving such decisions has been cross-entropy loss, which scales smoothly with compute and extrapolates with remarkable fidelity \citep{kaplan2020scalinglawsneurallanguage,hoffmann2022an}. 
However, the quantity we ultimately care about is downstream performance, not loss. Indeed, models with similar loss can exhibit sharply different downstream capabilities \citep{liu2023same}. Moreover, LLMs are increasingly judged on hard reasoning tasks where cross-entropy loss over generic text would offer little discriminative signal.


The natural response to resolve this discrepancy has been to fit scaling laws directly for downstream tasks, or to replace accuracy with smoother surrogates such as the likelihood of the correct answer \citep{gadre2025language,bhagia2025establishing,ruan2024observational,brandfonbrener2024l2l,hu2024predicting}. 
These approaches have been shown to work well when we assume access to plentiful evaluations on a target task, often with a closed answer set, or candidate models that perform above chance. 
However, the regimes in which downstream forecasting is most valuable are precisely those in which these assumptions are not met. 
Evaluations at the frontier of LLMs are often expensive or inaccessible, e.g., requiring human experts, code execution, or an external experimental loop \citep{patwardhan2025gdpval,wijk2025rebench}. 
Moreover, on hard reasoning tasks, small models or intermediate training checkpoints can all have indistinguishable accuracies \citep{hle}, which leaves no ordinal signal to fit. 
Recent work has also cast doubt on the reliability of downstream scaling laws themselves, finding that many task-level fits break when asked to extrapolate \citep{lourie2025unreliable}. 
The obstacle is not only that evaluation is expensive, but also that the quantities we can measure are often too sparse, too late, or too weakly tied to the reasoning process we hope to forecast.

\begin{figure}[t]
    \centering
    \begin{subfigure}[t]{0.48\textwidth}
        \includegraphics[width=\textwidth]{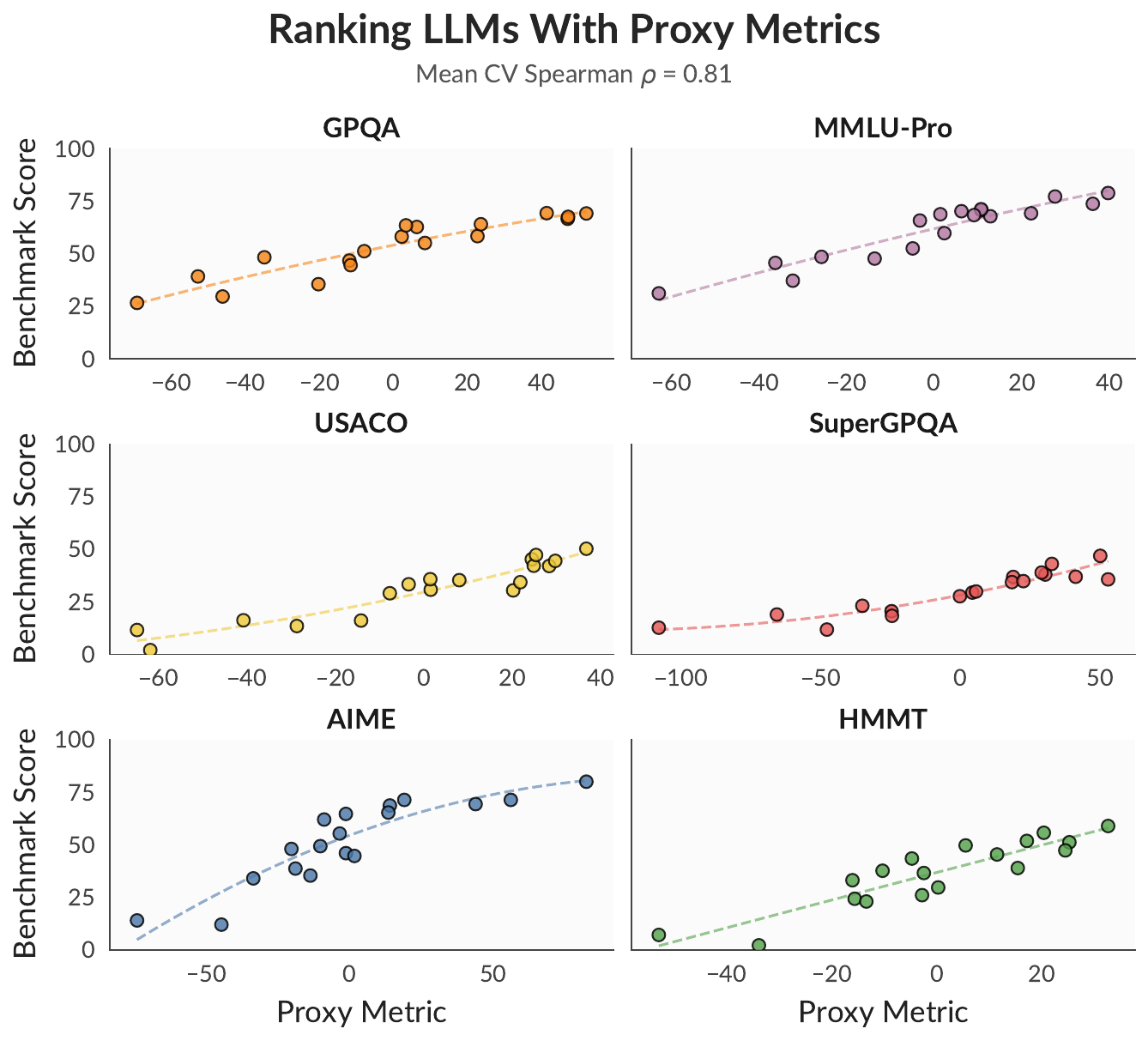}
    \end{subfigure}
    \hfill
    \begin{subfigure}[t]{0.51\textwidth}
        \includegraphics[width=\textwidth]{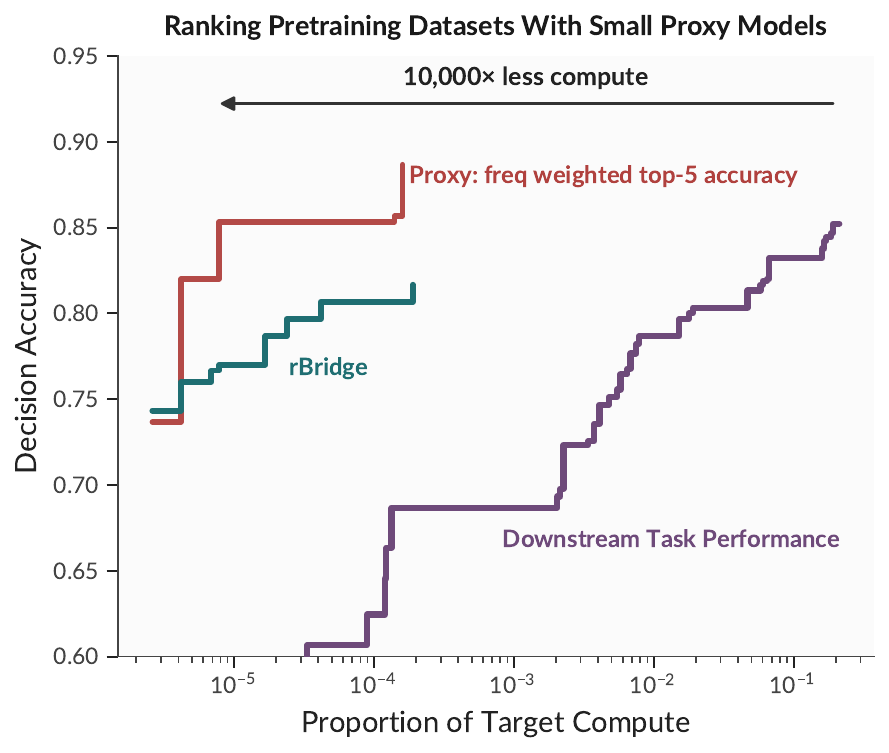}
    \end{subfigure}
    \caption{\textbf{Left.} Ranking models on held-out challenging reasoning tasks (as measured by mean CV Spearman $\rho$) using our linear RankSVM proxy. Our proxy uses features of the next-token prediction distributions of candidate models over expert reasoning traces. \textbf{Right:} Ranking 25 pretraining corpora for a target 1B LLM on the DataDecide testbed \citep{magnusson2025datadecide}. Each method trains small proxy models (4M--90M) on each corpus and attempts to recover the ground-truth corpus ranking defined by the target 1B model's downstream accuracy. Decision accuracy is the fraction of corpus pairs ranked correctly. Our best proxy metric pushes the Pareto frontier, requiring roughly $10{,}000\times$ less compute to match the downstream evaluation baseline.
    }
    \label{fig:main}
\end{figure}

In this paper, we propose a different approach for forecasting model performance: \textbf{compute proxy metrics based on the predictive distribution of the candidate model while it processes an expert solution.}
Our intuition is the following. 
A final benchmark score records only whether the model succeeded or failed, but an expert trajectory contains a long sequence of local decisions, and a model that cannot yet solve a task may still assign high probability to the crucial steps once they appear in context.
We build on this intuition by passing expert-written trajectories through the candidate in a single forward pass\footnote{Our approach does not require generating from the candidate model, and hence is extremely efficient.} and computing \emph{token-level} statistics of its next-token distribution, e.g., entropy, top-$k$ accuracy, rank of the expert token, etc.
These statistics are aggregated with weights that emphasize important positions, such as rare tokens or tokens where the candidate is uncertain.
Crucially, because the expert need only provide text, the same construction can use human solutions or traces from closed-weight frontier models.

We demonstrate our approach across three settings that mirror practical decisions in model development (\cref{fig:main}).
\circled{1} In \textbf{cross-family model selection} (\S\ref{sec:ranking_llms}), where the goal is to rank heterogeneous models on a downstream task without direct evaluation, our best proxy ranks models on held-out reasoning benchmarks in close agreement with their true performance (mean Spearman $\rho = 0.81$, compared with $0.36$ for cross-entropy loss).
\circled{2} In \textbf{pretraining data selection} (\S\ref{sec:ranking_datasets}), where the goal is to choose among candidate corpora before committing target-scale compute, our proxies reliably rank 25 diverse corpora using only small proxy models, achieving the same ranking quality as direct downstream evaluation at roughly $10{,}000\times$ less compute.
\circled{3} In \textbf{training-time forecasting} (\S\ref{sec:extrapolation}), we show that proxy metrics follow smooth power laws along training trajectories, enabling extrapolation from early checkpoints, and that downstream accuracy is more predictable as a function of our proxy metric compared to cross-entropy loss or compute, roughly halving extrapolation error across an $18\times$ compute horizon. The pattern across all settings is the same: generic loss is smooth but task-agnostic, direct evaluation is task-specific but expensive and often uninformative at early training stages, and expert-trajectory proxies provide both smoothness and task-conditioning in a single forward pass.

\section{Related Work}
\label{sec:related}

\paragraph{Scaling laws and downstream forecasting.}

Classical scaling laws predict pretraining loss as a function of compute, parameters, and data \citep{kaplan2020scalinglawsneurallanguage, hoffmann2022an}. 
Subsequent work has attempted to extend this predictability to downstream task performance, whether by fitting accuracy directly against compute \citep{owen2024predictablelanguagemodelbenchmark, krajewski2026revisiting}, mapping validation perplexity to downstream error \citep{gadre2025language}, decomposing the prediction into a compute-to-task-loss and task-loss-to-accuracy pipeline \citep{bhagia2025establishing}, fitting a latent capability axis over benchmark scores from public models \citep{ruan2024observational}, or linking loss thresholds to capability emergence \citep{du2024loss_emergence}. 
However, these approaches rest on assumptions that are often unmet in practice. 
Most require either a family of models trained at multiple scales or non-trivial benchmark scores across a broad population, neither of which is available when evaluating a new architecture or a single training run, or a task whose environment is inaccessible. 
Moreover, \citet{lourie2025unreliable} find that only a minority of downstream scaling laws extrapolate reliably, and \citet{liu2023same} demonstrate that models with nearly identical loss can differ substantially in downstream performance. 
A separate line of work predicts smoother task-specific losses across distributions \citep{brandfonbrener2024l2l, mayilvahanan2025llms}, avoiding the brittleness of accuracy, but this requires a closed answer set and does not resolve whether task loss tracks the downstream performance we ultimately care about. 
In this work, we focus on the problem of relative model ranking and show that proxy metrics derived from a single forward pass over expert trajectories can rank models on held-out tasks and across unseen models. 
Moreover, the benchmarks we consider such as graduate-level science \citep{rein2023gpqa} and olympiad programming \citep{shi2024usaco}, are precisely the reasoning tasks on which prior downstream scaling law approaches have not been tested.

\paragraph{Small-scale proxies for pretraining decisions.}

A separate line of work asks whether small proxy models can rank candidate pretraining corpora before committing target-scale compute. 
Prior approaches have selected domain weights \citep{xie2023doremi} or data mixtures \citep{liu2025regmix} using cheap small-scale runs. \citet{magnusson2025datadecide} systematize this question with DataDecide, a controlled testbed of twenty-five pretraining corpora at fourteen proxy scales, and show that likelihood-style metrics predict the 1B target ranking at $0.01\%$ of target compute. 
\citet{koh2025rbridge} improve on this with rBridge, which reweights the proxy model's likelihood by the expert model's token-level probabilities, defining the previous state-of-the-art Pareto frontier on DataDecide. 
Our proxy metrics displace this frontier while requiring only the expert's tokens, not its probabilities, which opens the door to closed-weight models and human experts as sources of expert signal. 
An extended discussion of other related works is provided in Appendix \ref{app:extended_related}.



\begin{figure}[t]
\centering
\includegraphics[width=\textwidth]{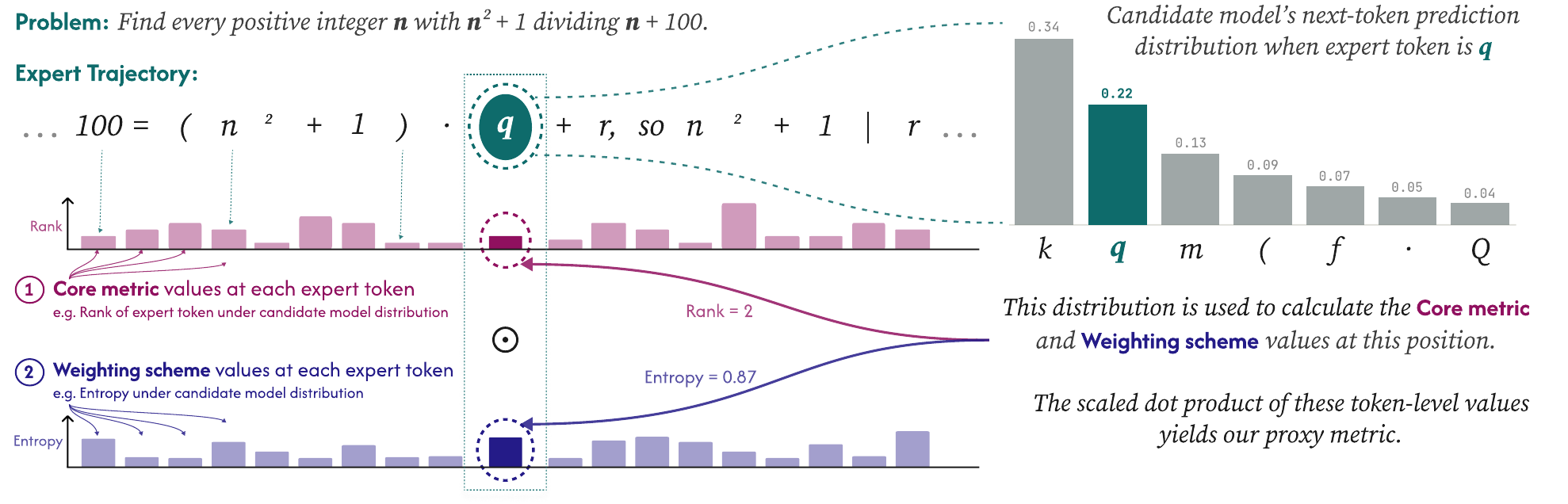}
\caption{An illustration of our method. We use a candidate model's next-token prediction distribution at each token of an expert's trajectory to calculate our proxy metrics.}
\label{fig:method_fig}
\end{figure}

\section{Method}
\label{sec:method}

Our goal is to design a proxy signal that is both indicative of a candidate model's capability on a downstream task, and cheap to evaluate.
We construct this signal from the candidate's predictive distribution over expert reasoning trajectories for the task instances as illustrated in \Cref{fig:method_fig}.
The intuition is that a model whose distribution often matches the expert's reasoning at every step is one that has internalized how the task is solved, even when its own generation might have failed. 
We assume access to such expert trajectories, whether written by humans or by strong language models. 
Reference solutions are already standard for benchmarks of practical interest, and for frontier domains where current LLMs are not yet competent, e.g., drug discovery, protein design, or theorem proving, domain experts working in tandem with AI can provide high-quality reasoning traces. 

\paragraph{Preliminaries.}

\begin{table}[t]
  \centering
  \footnotesize
  \caption{Core metrics (left) and weighting schemes (right) constituting our proxy metric library. We write $p \coloneqq p_\model(\cdot \mid \promptx^{(i)}, \traj^{(i)}_{<t})$, and $y_t$ refers to the expert token. See Appendix \ref{app:proxy_details} for details.}
  \label{tab:proxy_components}
  \begin{subtable}[t]{0.48\textwidth}
    \centering
    \caption{Core metrics}
    \label{tab:core_metrics}
    {\renewcommand{\arraystretch}{1.4}%
    \begin{tabular}{
      >{\raggedright\arraybackslash}m{9em}
      >{\centering\arraybackslash}m{12em}
      }
      \toprule
      \textbf{Core Metric} ($m_t$) & \textbf{Definition} \\
      \midrule
      Cross-entropy ($\mathcal{L}_t$) & $-\log p(y_t)$ \\
      \hdashline[0.5pt/2pt]
      Top-$k$ accuracy & $\mathbf{1}[y_t \in \mathrm{top\text{-}}k]$ \\
      & $k \in \{1,2,3,5\}$ \\
      \hdashline[0.5pt/2pt]
      Entropy ($H_t$) & $-\!\sum_v p(v)\log p(v)/\log|\mathcal{V}|$ \\
      \hdashline[0.5pt/2pt]
      Rank & $\mathrm{rank}(y_t)$ \\
      \hdashline[0.5pt/2pt]
      Reciprocal rank & $1/\mathrm{rank}(y_t)$ \\
      \hdashline[0.5pt/2pt]
      Margin & $\max_v p(v) - p(y_t)$ \\
      \hdashline[0.5pt/2pt]
      Wrong-confidence & $\max_v p(v)\!\cdot\!\mathbf{1}[\mathrm{rank}(y_t)\!>\!1]$ \\
      \bottomrule
    \end{tabular}}
  \end{subtable}%
  \hfill
  \begin{subtable}[t]{0.48\textwidth}
    \centering
    \caption{Weighting schemes}
    \label{tab:weights}
    {\renewcommand{\arraystretch}{1.4}%
    \begin{tabular}{
      >{\raggedright\arraybackslash}m{9em}
      >{\centering\arraybackslash}m{11em}}
      \toprule
      \textbf{Scheme} ($w_t$) & \textbf{Definition} \\
      \midrule
      Uniform & $1$ \\
      \hdashline[0.5pt/2pt]
      Probability & $p(y_t)$ \\
      \hdashline[0.5pt/2pt]
      Expert-disagreement & $1 - p(y_t)$ \\
      \hdashline[0.5pt/2pt]
      Entropy & $H_t$ \\
      \hdashline[0.5pt/2pt]
      Inverse entropy & $1 - H_t$ \\
      \hdashline[0.5pt/2pt]
      Frequency & $\mathrm{freq}(y_t)$ \\
      \hdashline[0.5pt/2pt]
      Inverse frequency & $1 - \mathrm{freq}(y_t)$ \\
      \hdashline[0.5pt/2pt]
      Gaussian-NLL & $e^{-(\mathcal{L}_t - \bar{\mathcal{L}})^2 / 2\sigma_{\mathcal{L}}^2}$ \\
      \bottomrule
    \end{tabular}}
  \end{subtable}
\end{table}

Given a downstream task $\task$ with instances $\{\promptx^{(i)}\}_{i=1}^{N}$ and
expert trajectories $\{\traj^{(i)}\}_{i=1}^{N}$, we pass each
$(\promptx^{(i)}, \traj^{(i)})$ pair through the candidate model $\model$. At each token position $t$ we obtain the predictive
distribution $p_\model(\cdot \mid \promptx^{(i)}, \traj^{(i)}_{<t})$, from which
we calculate a set of \emph{core metrics} $m_t$ and \emph{weighting schemes} $w_t$.
The 10 core metrics (\Cref{tab:core_metrics}) span three aspects of model--expert
alignment: how often the model agrees with the expert, how concentrated its
distribution is, and how confidently it errs when it disagrees.
Because not every token position is equally diagnostic, we aggregate each core
metric as a weighted average under eight weighting schemes (\Cref{tab:weights})
that emphasize different notions of token importance such as model uncertainty,
disagreement with the expert, or token rarity.

\paragraph{Proxy metrics.}
Each (metric, weighting) pair defines one \emph{proxy metric}, indexed by $j$.
Given an instance $(\promptx^{(i)}, \traj^{(i)})$ with trajectory of length
$T^{(i)}$, the proxy metric value is
\begin{equation}
\Phi_j(\model; \promptx^{(i)}, \traj^{(i)}) = \frac{\sum_{t=1}^{T^{(i)}} s_j \cdot m_{j,t}^{(i)} \cdot w_{j,t}^{(i)}}{\sum_{t=1}^{T^{(i)}} w_{j,t}^{(i)}},
\label{eq:proxy}
\end{equation}
where $m_{j,t}^{(i)}$ is the core metric value and $w_{j,t}^{(i)}$ is the weighting scheme value, both determined by $j$, at position $t$ of instance $i$, and $s_j \in \{+1, -1\}$ is a sign convention so that higher values indicate a better model (e.g., $s_j = -1$ for cross-entropy loss). The task-level proxy metric is the mean over instances,
\begin{equation}
\Phi_j(\model, \task) = \frac{1}{N} \sum_{i=1}^{N} \Phi_j(\model; \promptx^{(i)}, \traj^{(i)}).
\label{eq:proxy_task}
\end{equation}
With 10 core metrics and 8 weightings, we obtain a library of 80 proxy metrics $\Phi_1, \ldots, \Phi_{80}$, each assigning a scalar score $\Phi_j(\model, \task)$ to a candidate model on a task. When needed, we write $\Phi(\model, \task) \in \mathbb{R}^{80}$ for the full vector. The entire library is extracted from a single forward pass per instance, making computation extremely cheap while providing 80 complementary views of how closely the candidate's predictive distribution tracks the expert's reasoning.


\paragraph{Computing proxy metrics in practice.}
In all experiments we compute proxy metrics on the last 1{,}000 tokens of each expert trajectory, which empirically outperforms using the full trace. We do not filter out trajectories that yield incorrect answers, simulating the realistic setting of imperfect experts. When multiple experts are available, the 80 proxy metrics are averaged across experts and across instances, yielding $\Phi(\model, \task) \in \mathbb{R}^{80}$ per (model, task) pair.


\section{Cross-Family Model Selection: Ranking LLMs on Unseen Tasks}
\label{sec:ranking_llms}

A recurring decision in LLM development is choosing which of several candidate models will perform best on a downstream task of interest. 
The candidates may span different architectures, pretraining corpora, or post-training recipes, and the target evaluation is often inaccessible, requiring expert graders, code execution, or domain-specific infrastructure \citep{patwardhan2025gdpval, wijk2025rebench} that cannot be assembled at decision time. 
In this section we study whether the proxy metrics from \S\ref{sec:method} can be used to rank a heterogeneous model population on downstream tasks.



\subsection{Experimental Setup}


We evaluate 18 reasoning-capable language models spanning six model families and six post-training recipes, with sizes ranging from 0.6B to 70B parameters (full list in Appendix \ref{app:models}), on six challenging reasoning benchmarks: \textbf{AIME} 2025 \citep{aime25}, \textbf{HMMT} \citep{balunovic2025matharena}, \textbf{GPQA} \citep{rein2023gpqa}, \textbf{USACO} \citep{shi2024usaco}, \textbf{MMLU-Pro} \citep{wang2024mmlupro}, and \textbf{SuperGPQA} \citep{supergpqa} (details provided in Appendix \ref{app:benchmarks}). Together these cover competition math, graduate-level science, broad professional knowledge, and competitive code. Expert trajectories are generated by three frontier open-weight reasoning models: Kimi-K2.5 \citep{kimiteam2026kimik25visualagentic}, MiniMax-M2.5 \citep{minimax}, and Qwen3-Next-80B \citep{qwen3technicalreport}. We measure ranking quality by Spearman rank correlation ($\rho$) between proxy scores and downstream accuracy.

A natural first question is whether any single proxy metric is universally predictive across these tasks. To investigate, we select the best proxy using downstream scores from all six benchmarks and the full model population, an oracle setting that upper-bounds what any selection procedure can achieve (\cref{tab:oracle_results,tab:oracle_proxies} in the Appendix). The best proxy metric attains a mean $\rho$ of $0.62$, with per-task correlations ranging from $0.43$ to $0.81$. No single metric dominates universally. A linear combination of just three proxy metrics, however, reaches $\rho = 0.88$, indicating that the signal is present in the library but distributed across complementary metrics. A practitioner, however, will not have scores on the target task. We therefore ask: \textit{given downstream accuracy on a subset of tasks and models, can we find a proxy that generalizes to held-out tasks and unseen models?}


\paragraph{Evaluation protocol.}

We use a two-level resampling scheme.
At the \emph{task} level, we perform leave-2-tasks-out cross validation over the six benchmarks, producing $\binom{6}{2}=15$ folds.
In each fold the proxy is selected on the four held-in tasks and scored by the mean $\rho$ on the two held-out tasks.
At the \emph{model} level, for each fold we further sample $60\%$ of the models at random for selection and evaluate ranking correlation on the full model set.
We repeat the model sampling with 20 fixed seeds and report mean $\pm$ std across seeds.

\paragraph{Ranking models.}

The 80 proxy metrics from \S\ref{sec:method} reduce ranking to a low-dimensional learning problem: we seek a function $f : \mathbb{R}^{80} \to \mathbb{R}$ whose induced ordering over candidate models tracks their downstream ordering. We compare four model classes of increasing capacity: a \textbf{univariate proxy} $f(\Phi) = \Phi_j$; a \textbf{$3$-sparse proxy} $f(\Phi) = \sum_{k=1}^{3} \alpha_k \Phi_{j_k}$; a \textbf{linear RankSVM} $f(\Phi) = \mathbf{w}^\top \Phi$ trained under a pairwise hinge loss \citep{herbrichranksvm}; and an \textbf{RBF RankSVM} $f(\Phi) = \sum_i \alpha_i \, k(\Phi, \Phi_i)$ with a Gaussian kernel, trained under the same objective.

\paragraph{Proxy selection.}
For the univariate proxy, we select the index $j$ that maximizes the mean Spearman $\rho$ between $\Phi_j(\cdot, \task)$ and downstream accuracy, averaged over tasks $\task \in \mathcal{T}_{\text{in}}$.
For the $3$-sparse proxy, we enumerate all $\binom{80}{3}$ index triplets and sweep a signed log-spaced grid of coefficient ratios in $[-10^3, 10^3]$, selecting the triplet and ratios that maximize the same objective.
For both RankSVM variants, the parameters are fit on $(\model_i \succ \model_j)$ preference pairs induced by the downstream scores.

\paragraph{Baselines.}

Cross-entropy loss on generic text has been widely used as a predictor of downstream capability \citep{du2024loss_emergence, brandfonbrener2024l2l, mayilvahanan2025llms}.
We compute CE loss over 10M tokens from randomly sampled FineWeb \citep{penedo2024the} documents.
We also evaluate rBridge \citep{koh2025rbridge}, which computes expert-probability-weighted CE loss over expert reasoning chains, requiring access to the expert model's logprobs.

\subsection{Results and Discussion}

\begin{table}[t]
    \centering
    \caption{Leave-2-tasks-out cross-validated Spearman rank correlation ($\rho$) between each proxy and downstream benchmark accuracy. The per-task columns report the mean test $|\rho|$ across all folds in which that task was held out, and the \textbf{Mean $\rho$} column averages over all folds. For learned proxies, values are additionally averaged over 20 random model-subsampling seeds and reported as mean\,$\pm$\,std across seeds. Best per column is shown in \textbf{bold}.}
    \label{tab:main_results}
    \setlength{\tabcolsep}{4pt}
    \small
    \newcommand{\stdpm}[1]{{\scriptsize\textcolor{gray!70!black}{$\pm$#1}}}
    {\renewcommand{\arraystretch}{1.15}%
    \begin{tabular}{l c cccccc}
        \toprule
        \textbf{Proxy} & \textbf{Mean $\rho$} & AIME & GPQA & HMMT & MMLU-Pro & SuperGPQA & USACO \\
        \midrule
        FineWeb CE loss        & 0.36 & 0.37 & 0.25 & 0.25 & 0.52 & 0.54 & 0.25 \\
        rBridge & 0.33 & 0.22 & 0.37 & 0.13 & 0.33 & 0.55 & 0.39 \\
        \midrule
        Univariate proxy & 0.54\stdpm{0.08} & 0.44\stdpm{0.09} & 0.57\stdpm{0.05} & 0.42\stdpm{0.04} & 0.62\stdpm{0.12} & 0.64\stdpm{0.17} & 0.53\stdpm{0.13} \\
        $3$-sparse proxy & 0.78\stdpm{0.06} & 0.74\stdpm{0.09} & 0.86\stdpm{0.07} & 0.68\stdpm{0.11} & 0.82\stdpm{0.06} & 0.84\stdpm{0.07} & 0.73\stdpm{0.08} \\
        RankSVM (RBF)          & \textbf{0.81}\stdpm{0.05} & \textbf{0.75}\stdpm{0.11} & 0.84\stdpm{0.09} & \textbf{0.76}\stdpm{0.09} & 0.87\stdpm{0.07} & 0.84\stdpm{0.07} & \textbf{0.78}\stdpm{0.09} \\
        RankSVM (linear)       & \textbf{0.81}\stdpm{0.04} & 0.73\stdpm{0.11} & \textbf{0.87}\stdpm{0.07} & 0.73\stdpm{0.10} & \textbf{0.88}\stdpm{0.07} & \textbf{0.88}\stdpm{0.05} & \textbf{0.78}\stdpm{0.10} \\
        \bottomrule
    \end{tabular}}
\end{table}

\Cref{tab:main_results} reports the leave-2-tasks-out Spearman $\rho$ for all four proxy models and the two loss-based baselines, aggregated across 15 folds and 20 model-subsampling seeds.

\paragraph{Loss-based baselines fail to rank models.}

CE loss on FineWeb achieves only $\rho = 0.36$, confirming that a scalar summary of fit to generic text carries little information about relative performance on reasoning tasks. rBridge, which reweights the likelihood along a frontier-model reasoning trace and has access to expert logprobs, fares no better at $\rho = 0.33$. These results are further illustrated in \Cref{fig:unseen_scatter} (left) in the Appendix, where we visualize the loss-based baselines against MMLU-Pro accuracy and find no coherent pattern.

\paragraph{Proxy models show high correlation with performance.}

The univariate proxy reaches $\rho = 0.54$, which is higher than the best loss-based baseline. 
The $3$-sparse proxy pushes this to $\rho = 0.78$, and the full linear RankSVM reaches $\rho = 0.81$, with the RBF variant tied. 
\Cref{fig:main} (left) plots downstream accuracy against the linear RankSVM proxy score for each of the six benchmarks in a randomly sampled held-out fold. 
Across all six tasks the relationship is monotonic. 
\Cref{fig:unseen_scatter} (right) in the Appendix zooms into MMLU-Pro, showing that this monotonic relationship holds across different base families and post-training recipes. \Cref{fig:sweep_held_perc_linear,fig:sweep_held_perc_sparse} in the Appendix show that similar trends hold even when we consider 3 held-out tasks and a lesser percentage of models used for selection.

\paragraph{Ranking signal concentrates on a few proxy metrics.}
\Cref{fig:feature_heatmap} in the Appendix shows how often each proxy metric is selected across folds and seeds. The distribution concentrates on a handful of cells.
For the univariate proxy, inverse-frequency-weighted top-1 accuracy, a signal indicating model agreement with expert on rare tokens, dominates.
For the $3$-sparse proxy, entropy-weighted entropy and frequency-weighted top-5 accuracy are most frequently selected, capturing model uncertainty at positions where the candidate is least committed.
These are precisely the ``forking tokens'' that \citet{wang20258020rulehighentropyminority} identify as driving the majority of RL training signal in reasoning chains.
Note that this analysis characterizes where the ranking signal concentrates rather than explaining why the proxy works, which does not depend on the selected features being interpretable.



\section{Pretraining Data Selection: Ranking Datasets with Smaller LMs}
\label{sec:ranking_datasets}

Before committing to a target-scale pretraining run, a team must choose among various candidate pretraining corpora. The target run may cost millions of dollars, so the decision should ideally rest on evidence collected at a fraction of that budget. The standard approach is to train small proxy models on each candidate corpus and rank them by downstream benchmark accuracy or cross-entropy loss \citep{magnusson2025datadecide}. But at small scale, benchmark accuracy is noisy or at chance, and cross-entropy loss on generic text correlates poorly with downstream performance. 
In this section, we ask: \emph{can our proxy metrics, computed over small proxy models, rank pretraining corpora without ever evaluating on downstream tasks?}


\subsection{Experimental Setup}

We use the DataDecide testbed \citep{magnusson2025datadecide}, which consists of twenty-five candidate pretraining corpora, each used to train proxy models at scales ranging from 4M to 90M parameters, together with realized 1B-parameter target models trained on the same corpora. 
The ground-truth ranking of the twenty-five corpora is defined by the mean downstream accuracy of the corresponding 1B target models on the \textbf{OLMES} suite of ten multiple-choice benchmarks \citep{gu2024olmesstandardlanguagemodel}.


\paragraph{Evaluation metric.}

Following \citet{magnusson2025datadecide}, we measure ranking quality by \emph{decision accuracy}, which is the fraction of corpus pairs in which the proxy model agrees with the target-scale ranking. 
It can be formalized as follows. 
Let \(\mathcal{P}\) be the set of all pretraining corpus pairs \((A, B)\) with observed mean OLMES performance for the 1B target LLM as \(y_A, y_B\) respectively, and the predicted performance by the proxy model is denoted by \(\hat{y}_A, \hat{y}_B\), respectively, then decision accuracy is:

\begin{equation}
    \textstyle \frac{1}{\lvert \mathcal{P}\rvert} \sum_{(A, B) \in \mathcal{P}} \mathbb{I}\big(\text{sign}(\hat{y}_A - \hat{y}_B) = \text{sign}(y_A - y_B)\big)
\end{equation} 

\paragraph{Estimating compute.}

We measure the cost of a ranking method by the fraction of the 1B target's training FLOPs consumed by the proxy model, following the standard approximation $\text{FLOPs} = 6ND$ \citep{kaplan2020scalinglawsneurallanguage}. 
A method that ranks corpora using only 4M-parameter proxy models operates at roughly $10^{-5}$ of the target compute.

\paragraph{Method.}

We evaluate each univariate proxy metric on every (corpus, model-size) pair in DataDecide, producing a ranking of the $25$ corpora at each compute budget. 
Due to the simpler nature of the OLMES tasks compared to the tasks in \S\ref{sec:ranking_llms}, here we use chain-of-thought (CoTs) produced by Llama-3.3-70B \citep{grattafiori2024llama3herdmodels} as the expert trajectories for computing our proxy metrics. 
We compare against two baselines. 
The first is the DataDecide downstream task performance baseline, which scores each corpus by the mean OLMES accuracy of the corresponding proxy model. 
The second is rBridge \citep{koh2025rbridge}, which reweighs the proxy model's negative log-likelihood at each token position of the expert's CoT by the expert model's token-level probabilities. 



\subsection{Results and Discussion}

\Cref{fig:main} (right) plots decision accuracy against compute budget for all methods.
Prior to this work, rBridge defined the state-of-the-art Pareto frontier on DataDecide.

\paragraph{New state-of-the-art on DataDecide.}

Our best proxy metric, frequency-weighted top-5 accuracy, pushes this frontier.
At roughly $10^{-5}$ of the target compute, it reaches decision accuracy above $0.85$, and at matched compute budgets it dominates rBridge at almost every point where both methods are defined.
To reach comparable decision accuracy using the downstream performance baseline, one must scale proxy models to a budget exceeding $10^{-1}$ of the target, which is roughly $10{,}000$ times more compute.
This advantage comes under \emph{strictly weaker assumptions}: our proxy metrics require only the expert's discrete tokens, not its logprobs, enabling closed-weight frontier models as well as human experts to be valid sources.

\paragraph{Why proxies are suitable for small models.}
We believe that the reason proxy metrics discriminate among corpora at these scales is that a model which cannot solve a problem can still track the CoT 
written by an expert.
Benchmark accuracy requires the model to produce a correct answer, which may not be possible at such small scales.
But the model's token-level compatibility with an expert trajectory varies measurably across corpora long before any benchmark score exits the noise floor.

\section{Training-time Forecasting: Proxy Metrics Facilitate Extrapolation}
\label{sec:extrapolation}

The previous sections showed that proxy metrics can rank models and pretraining corpora for downstream task performance.
But a practitioner often needs to forecast end-of-training capability with a fraction of the training compute expended.
This requires that the proxy evolve predictably with training compute, so that a fit from early checkpoints can be extrapolated to later ones.

\subsection{Extrapolating Proxy Metric During Training}
\label{sec:extrap_proxy}

\begin{figure}[t]
\centering
\begin{subfigure}[t]{0.49\textwidth}
    \includegraphics[width=\textwidth]{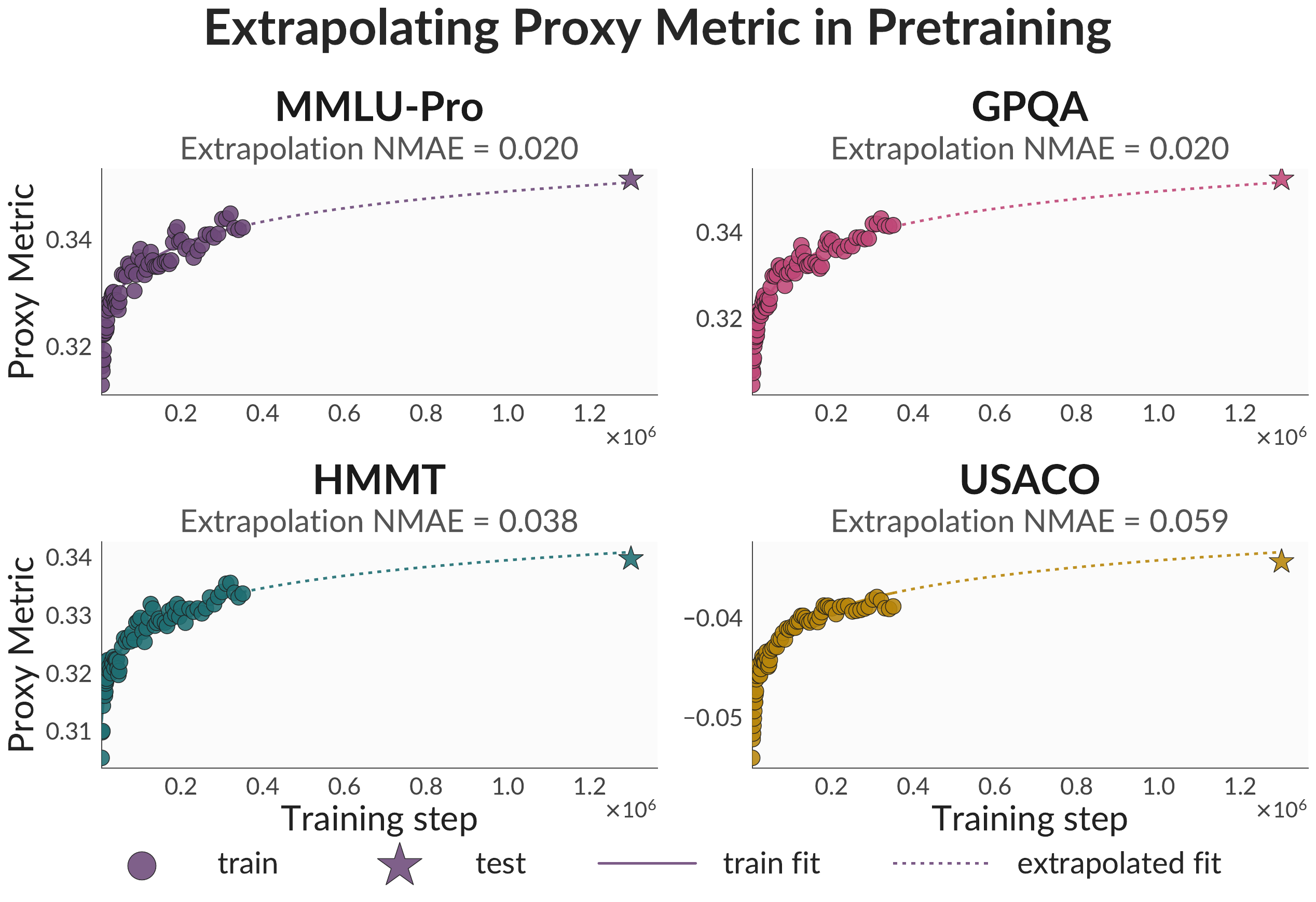}
\end{subfigure}
\hfill
\begin{subfigure}[t]{0.49\textwidth}
    \includegraphics[width=\textwidth]{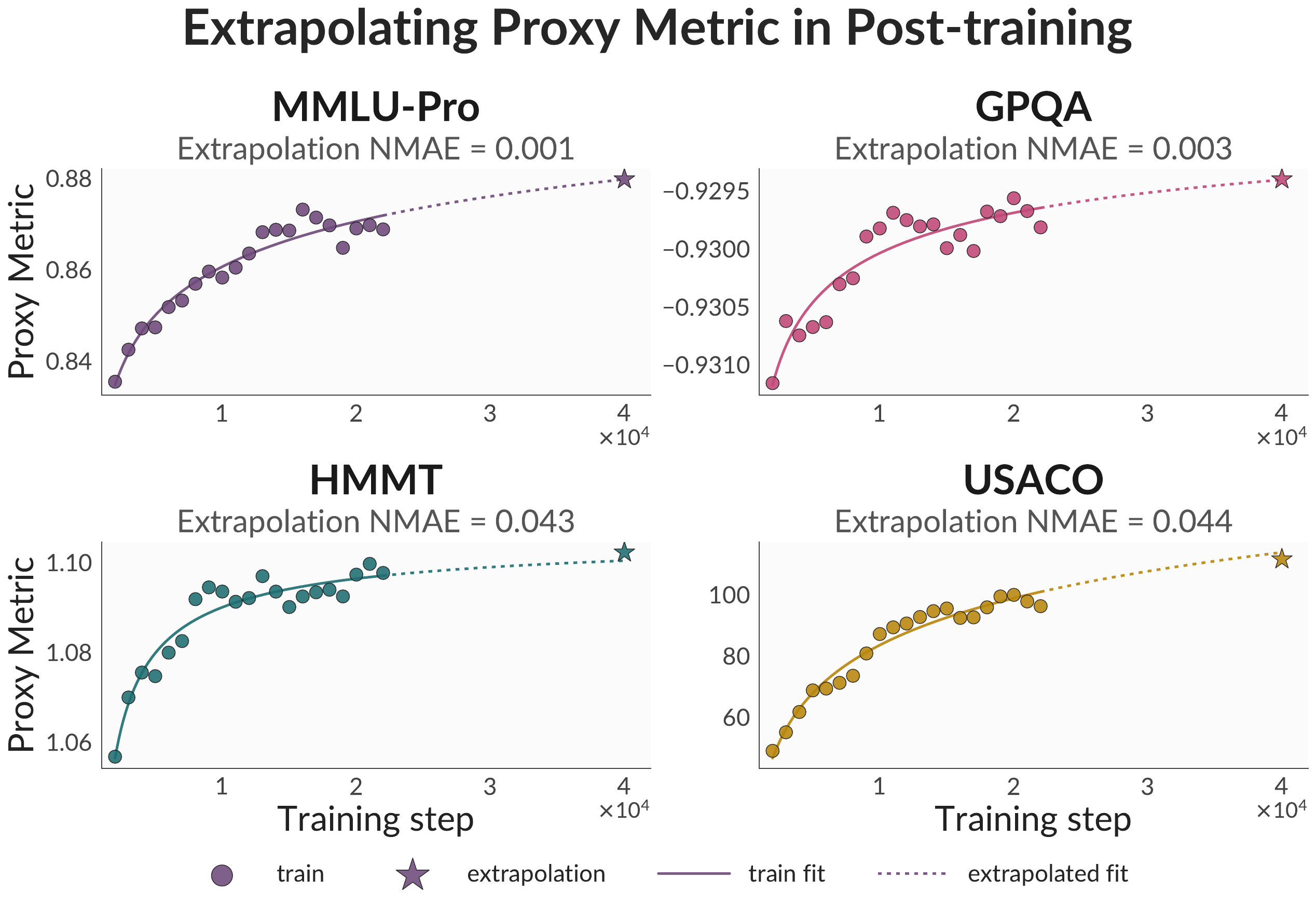}
\end{subfigure}
\caption{\textbf{Extrapolating proxy metrics along the training trajectory.} \emph{Left:} pretraining checkpoints of OLMo-3-7B on four reasoning benchmarks. \emph{Right:} post-training checkpoints of OLMo-3-7B-Think on four reasoning benchmarks. Filled markers are the training window, stars are held-out checkpoints, solid curves are power-law fits from the training window, and dashed curves are extrapolations. The plots for other benchmarks are provided in \cref{fig:extrapolation_pre_aime_supergpqa,fig:extrapolation_post_aime} in the Appendix.
}
\label{fig:extrapolation}
\end{figure}

First, we test whether \emph{we can find proxy metrics that follow a simple functional form along the training trajectory}.

\paragraph{Experimental setup.}

We study two trajectories: the pretraining checkpoints of OLMo-3-7B \citep{olmo2026olmo3}, evaluated on all six reasoning benchmarks from \S\ref{sec:ranking_llms}, and the post-training checkpoints of OLMo-3-7B-Think, evaluated on five (excluding SuperGPQA, on which the model shows negligible improvement).
At each checkpoint we compute univariate proxy metrics and fit a power law $f(t) = a - b\, t^{-c}$.
We select the proxy via an inner split: fit on the first $k=50\%$ of checkpoints, choose the proxy whose extrapolation to the remainder has the lowest normalized mean absolute error (NMAE), defined as mean absolute error divided by the proxy's range over training.
The selected proxy is then refit on the full window and evaluated on the held-out checkpoint.

\paragraph{Results and discussion.}

The results are shown in \Cref{fig:extrapolation}.
In pretraining, we fit on checkpoints up to 350K steps and extrapolate to ${\sim}4\times$ the compute.
On all six benchmarks the best selected proxy follows a smooth power law, with mean NMAE of $0.03$ across tasks.
In post-training, the proxy extrapolates to nearly $2\times$ the training compute with mean NMAE of $0.038$.
As a sanity check, the best selected proxies also correlate strongly with downstream accuracy at the post-training checkpoints (mean Spearman $\rho = 0.84$ illustrated in Appendix \cref{fig:extrapolation_post_scatter_usaco_hmmt}), confirming that the extrapolated quantity tracks the ranking we care about.


\subsection{Downstream Accuracy Is More Predictable from Proxy Metrics}
\label{sec:extrap_downstream}

Existing approaches for predicting downstream task performance fit exponential functions over validation loss \citep{gadre2025language} or sigmoids over log-compute \citep{owen2024predictablelanguagemodelbenchmark}, but \citet{lourie2025unreliable} show that the majority of such fits fail to extrapolate reliably.
Hence, we next ask: \emph{is downstream accuracy more predictable as a function of our proposed proxy metrics compared to cross-entropy loss or training compute?}

\begin{figure}[t]
    \centering
    \begin{minipage}[t]{0.49\textwidth}
        \centering
        \captionof{table}{Extrapolation RMSE for predicting downstream accuracy at $1.4$M steps from checkpoints up to $80$K steps (${\sim}18\times$). \textbf{Bold} marks the lowest RMSE per task.}
        \label{tab:proxy_vs_downstream_main}
        \vspace{0.3em}
        \setlength{\tabcolsep}{4pt}
        \renewcommand{\arraystretch}{1.1}
        \small
        \begin{tabular}{lccc}
        \toprule
        Task & Proxy & CE Loss & Compute \\
        \midrule
        HellaSwag      & \textbf{0.003} & 0.094 & 0.105 \\
        BoolQ          & 0.037 & 0.111 & \textbf{0.034} \\
        SocialIQA      & 0.016 & \textbf{0.011} & 0.025 \\
        Winogrande     & \textbf{0.010} & 0.078 & 0.018 \\
        ARC Challenge     & \textbf{0.022} & 0.131 & 0.071 \\
        ARC Easy       & 0.049 & \textbf{0.001} & 0.036 \\
        CommonsenseQA           & \textbf{0.004} & 0.023 & 0.129 \\
        MMLU           & \textbf{0.001} & 0.036 & 0.058 \\
        OpenBookQA           & 0.069 & 0.086 & \textbf{0.000} \\
        PIQA           & 0.029 & \textbf{0.025} & 0.068 \\
        \midrule
        \textbf{Mean}  & \textbf{0.024} & 0.059 & 0.055 \\
        \bottomrule
        \end{tabular}
    \end{minipage}
    \hfill
    \begin{minipage}[t]{0.49\textwidth}
        \centering
        \vspace{0pt}
        \includegraphics[width=0.96\linewidth, trim=0 0 0 3cm, clip]{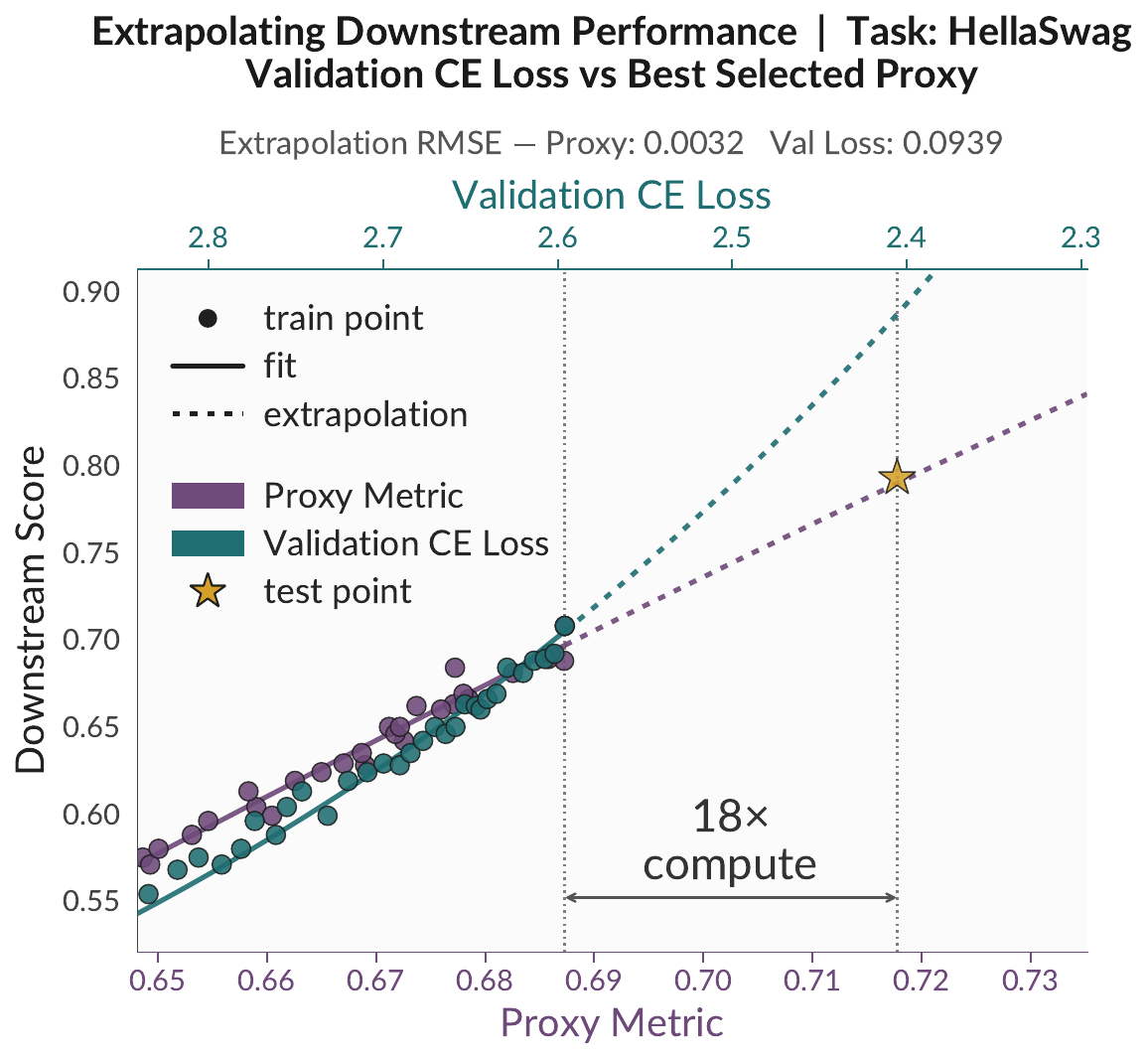}
        \caption{Extrapolating HellaSwag accuracy along the OLMo-3-7B pretraining trajectory. The proxy power-law fit (RMSE $= 0.003$) tracks the target far more closely than the CE loss exponential (RMSE $= 0.09$).}
        \label{fig:proxy_vs_downstream}
    \end{minipage}
\end{figure}

\paragraph{Experimental setup.}

We use pretraining checkpoints of OLMo-3-7B \citep{olmo2026olmo3} across ten OLMES benchmarks \citep{gu2024olmesstandardlanguagemodel}.
For each benchmark we fit a predictor$\to$accuracy curve on checkpoints up to $80{,}000$ steps and evaluate at the final checkpoint ($1.4$M steps, roughly $18\times$ the fitting horizon).
We compare three predictors: (1)~CE loss, via an exponential from FineWeb validation loss to accuracy \citep{gadre2025language}; (2)~compute, via a sigmoid against $\log_{10}(\text{steps})$ \citep{owen2024predictablelanguagemodelbenchmark}; and (3)~proxy metric, via a power law, selected by the inner-split protocol of \S\ref{sec:extrap_proxy}.


\paragraph{Results and discussion.}
\Cref{tab:proxy_vs_downstream_main} summarizes the extrapolation RMSE across all ten OLMES benchmarks.
The proxy-based fit achieves a mean RMSE of $0.024$, roughly half that of the CE loss predictor ($0.059$) and the compute-based predictor ($0.055$).
Even on the tasks where a baseline achieves the lowest RMSE, the proxy-based fit is comparable.
\Cref{fig:proxy_vs_downstream} illustrates the contrast on HellaSwag: within the training window all fits are strong ($R^2 > 0.9$), but at extrapolation the CE loss fit overshoots (RMSE $= 0.09$) while the proxy fit tracks the target closely (RMSE $= 0.003$).
We attribute this to the fact that CE loss and compute are both task-agnostic predictors, whereas the proxy is conditioned on the target task through the expert trajectories, and so its relationship with downstream accuracy is less likely to drift as training progresses.

\section{Conclusion}
\label{sec:conclusion}

We proposed proxy metrics computed from token-level statistics of a candidate model's forward pass over expert reasoning trajectories and showed that they carry substantial information about downstream task capability. 
\circled{1} For \textbf{cross-family model selection} under leave-two-tasks-out cross-validation across 6 reasoning benchmarks and 18 models spanning 6 base families, a linear ranker over these metrics achieves Spearman $\rho = 0.81$, compared to $\rho = 0.36$ for cross-entropy loss. 
\circled{2} For \textbf{pretraining data selection} on the DataDecide testbed, a single univariate proxy ranks 25 pretraining corpora with decision accuracy above $0.85$ at roughly $10^{-5}$ of target compute, displacing the prior Pareto frontier while requiring only the expert's tokens, not its logprobs. 
\circled{3} For \textbf{training-time forecasting} along the pretraining trajectory of OLMo-3-7B, proxy-to-accuracy fits extrapolate downstream performance across an $18\times$ compute horizon with roughly half the error of loss-based or compute-based alternatives. 
Across all three settings, the mechanism is the same: the candidate model's token-level distribution over an expert trajectory provides a dense, task-conditioned signal that inherits the smoothness of loss while remaining grounded in the reasoning process that the downstream evaluation is designed to measure.

Several important questions remain. Our experiments establish that proxy metrics work across model families and training stages, but the boundaries of this finding are not yet fully mapped: whether the same construction generalizes to mixture-of-experts architectures, to task types beyond reasoning such as long-context retrieval and agentic evaluation, and whether the power-law fits that enable extrapolation hold across model scales and not only along a single training trajectory. A related question concerns the expert itself. We have assumed access to high-quality reasoning traces, but in practice expert quality will vary, and understanding how proxy signal degrades with weaker or noisier experts is important for deployment. Finally, we have treated the eighty proxy metrics as a fixed library and selected among them post hoc. Learning the aggregation function end-to-end, or designing metrics that target specific failure modes of downstream evaluation, could close the remaining gap between the oracle upper bound and the cross-validated performance we report.

\section{Limitations}\label{app:limitations}

Our experimental scope has boundaries that should inform how the results are interpreted.

\paragraph{Model and checkpoint coverage.}
The cross-family experiment evaluates 18 models spanning six base families and six post-training recipes. This covers most of the major open-weight reasoning-capable models available while this work was done, but the population is still modest. The training-time forecasting experiments (\S\ref{sec:extrap_proxy}, \S\ref{sec:extrap_downstream}) rely entirely on checkpoints from OLMo-3-7B and OLMo-3-7B-Think. We chose OLMo because it publicly releases both pretraining and post-training checkpoints at the granularity needed for power-law fitting. This means the extrapolation results are established on a single architecture and scale, and whether the same power-law regularity holds across model sizes, architectures, or mixture-of-experts models remains an open empirical question.

\paragraph{No universal proxy metric.}
No single proxy metric dominates across all settings. The best univariate proxy for cross-family ranking is not the same as the best proxy for data selection or for extrapolating any particular benchmark. In practice, the specific proxy metric that works best will depend on the available held-in benchmarks, the model population, and hyperparameters of the selection procedure such as the inner-split fraction and the training-window size used for fitting. We have not conducted a systematic sensitivity analysis over these choices. The oracle gap (\cref{tab:oracle_results}) shows that the 80-dimensional library contains enough signal to reach $\rho = 0.88$ with just three features, but extracting that signal reliably under realistic selection constraints is challenging.

\paragraph{Task coverage.}
The cross-family ranking experiment focuses on challenging reasoning benchmarks, while the data selection and downstream extrapolation experiments use OLMES, a suite of non-reasoning multiple-choice tasks. This split is driven by resource constraints: DataDecide provides a controlled testbed only for OLMES, and, to the best of our knowledge, no comparable infrastructure exists for reasoning benchmarks at the time of writing. As a result, we have not demonstrated pretraining data selection for reasoning tasks or downstream extrapolation on hard reasoning benchmarks. More broadly, we have also not tested the method on generative tasks (e.g., open-ended writing, translation), long-context benchmarks, or agentic evaluations.

\section*{Acknowledgments}

\noindent Arkil is partly supported by the Canada Graduate Scholarships (Doctoral) funded by the Natural Sciences and Engineering Research Council (NSERC) [funding reference no. 601601]. 
We thank our colleagues at Mila and McGill University for helpful discussions and for providing valuable feedback.

\clearpage
\newpage

\bibliography{references}
\bibliographystyle{plainnat}

\newpage


\appendix

\section{Details of Experiments}\label{app:exp_details}

\subsection{Proxy metric definitions}\label{app:proxy_details}

Below, we describe the exact formulation of each of our core metrics and weighting schemes.

\paragraph{Core metrics.}

At each position $t$ we compute the \textbf{(1) cross-entropy loss} $\mathcal{L}_{t} = -\log p_\model(y_t\mid\promptx, y_{<t})$, \textbf{(2) top-$k$ accuracy} $A_{k,t} = \mathbf{1}[y_t \in \mathrm{top\text{-}}k(\model, t)]$ for $k \in \{1,2,3,5\}$, \textbf{(3) entropy} $H_t = -\sum_v p_\model(v\mid\promptx, y_{<t})\log p_\model(v\mid\promptx, y_{<t})/\log|\mathcal{V}|$ (vocabulary-normalized), \textbf{(4) rank} of $y_t$, \textbf{(5) reciprocal rank} $1/\mathrm{rank}(y_t)$, \textbf{(6) margin} $\max_v p_\model(v\mid\promptx, y_{<t}) - p_\model(y_t\mid\promptx, y_{<t})$, and \textbf{(7) wrong-confidence mass} $\max_v p_\model(v\mid\promptx, y_{<t}) \cdot \mathbf{1}[\mathrm{rank}(y_t) > 1]$ that the model places on the wrong token when it misses. Together these metrics span three regimes of model-expert alignment: how often the model agrees with the expert, how peaked the model is, and how badly the model errs when it errs.

\paragraph{Weighting schemes.}

Not every token position is equally important, e.g.,  
function words and punctuation are predicted well by most models and drown out signal. 
We therefore introduce per-token weights $w_t \geq 0$ and aggregate each metric as a weighted average. The eight schemes we consider are: \textbf{(1) uniform} ($w_t = 1$), \textbf{(2) probability} ($w_t = p_\model(y_t)$), \textbf{(3) expert-disagreement} ($w_t = 1-p_\model(y_t)$), \textbf{(4) entropy} ($w_t = H_t$, where $H_t$ is vocabulary-normalized so $H_t \in [0,1]$), \textbf{(5) inverse entropy} ($w_t = 1 - H_t$), \textbf{(6) frequency} ($w_t = \mathrm{freq}(y_t)$, where $\mathrm{freq}(y_t)$ is the frequency of the token in the entire expert trajectories corpus $\{(\promptx_i, \traj_i)\}_{i=1}^{N}$), \textbf{(7) inverse frequency} ($w_t = 1 - \mathrm{freq}(y_t)$), and \textbf{(8) Gaussian-NLL} kernel that isolates positions near the typical loss level ($w_t = \exp(-(\mathcal{L}_t - \bar{\mathcal{L}})^2/2\sigma_{\mathcal{L}}^2)$, where $\bar{\mathcal{L}}$ and $\sigma_\mathcal{L}$ are the mean and standard deviation of $\mathcal{L}_t$ along the trajectory).

\subsection{Model details}
\label{app:models}

We evaluate 18 reasoning-capable language models spanning six base families (Qwen3 \citep{qwen3technicalreport}, Qwen2.5 \citep{qwen2.5-1m}, Llama3 \citep{grattafiori2024llama3herdmodels}, Ministral \citep{liu2026ministral3}, OLMo3 \citep{olmo2026olmo3}, SmolLM3 \citep{bakouch2025smollm3}) and six post-training recipes (Qwen3, R1-distillation \citep{Guo_2025}, Nemotron \citep{wang2026nemotroncascadescalingcascadedreinforcement}, Ministral-Reasoning, OLMo3-Think, and SmolLM), with sizes varying from 0.6B to 70B parameters. \Cref{tab:model_list} lists these 18 models. We treat Qwen3-A3B as a part of the Qwen3 family. We additionally experimented with Ministral-3-14B-Reasoning, but could not obtain consistent performance evaluations on benchmarks (for instance, on some benchmarks it would perform even worse than the 3B model). We consider this a failure of our implementation or lack of reproducibility and hence do not include those results in our analysis.

\begin{table}[b]
\centering
\caption{Models evaluated in the cross-family rank forecasting experiment.}
\label{tab:model_list}
\smallskip
\small
\begin{tabular}{llll}
\toprule
\textbf{Base Family} & \textbf{Model} & \textbf{Post-training} & \textbf{Short Name} \\
\midrule
Qwen2.5 & R1-Distill-Qwen-7B & R1 & R1-Q-7B \\
Qwen2.5 & R1-Distill-Qwen-14B & R1 & R1-Q-14B \\
Qwen2.5 & R1-Distill-Qwen-32B & R1 & R1-Q-32B \\
Llama3 & R1-Distill-Llama-8B & R1 & R1-L-8B \\
Llama3 & R1-Distill-Llama-70B & R1 & R1-L-70B \\
Ministral & Ministral-3-3B-Reasoning & Ministral & Ministral-3B \\
Ministral & Ministral-3-8B-Reasoning & Ministral & Ministral-8B \\
OLMo3 & OLMo-3-32B-Think & OLMo3 & OLMo3-32B \\
SmolLM3 & SmolLM3-3B & SmolLM & SmolLM3-3B \\
Qwen3 & Qwen3-0.6B & Qwen3 & Qwen3-0.6B \\
Qwen3 & Qwen3-1.7B & Qwen3 & Qwen3-1.7B \\
Qwen3 & Qwen3-4B & Qwen3 & Qwen3-4B \\
Qwen3 & Qwen3-8B & Qwen3 & Qwen3-8B \\
Qwen3 & Qwen3-14B & Qwen3 & Qwen3-14B \\
Qwen3 & Qwen3-32B & Qwen3 & Qwen3-32B \\
Qwen3-A3B & Qwen3-30B-A3B-Thinking & Qwen3-A3B & Qwen3-A3B-30B \\
Qwen3 & Nemotron-Cascade-8B & Nemotron & Nemo-8B \\
Qwen3 & Nemotron-Cascade-14B & Nemotron & Nemo-14B \\
\bottomrule
\end{tabular}
\end{table}

We limit generations for all models to a maximum of 28000 tokens. This is to ensure fair comparison across models since some support shorter context lengths than others. In general, we generate with temperature $T=0.6$, $\text{top-}p = 0.95$, and $\text{top-}k$ disabled. Where needed, we followed model-specific overrides based on developer recommendations, for e.g., Ministral $\le$ 8B uses $T=0.7$, along with a special system prompt.\footnote{\href{https://huggingface.co/mistralai/Ministral-3-8B-Reasoning-2512}{https://huggingface.co/mistralai/Ministral-3-8B-Reasoning-2512}} All benchmark performance evaluations are averaged over 5 random seeds.

\subsection{Details of Benchmarks}\label{app:benchmarks}

In our experiments in \cref{sec:ranking_llms} and \cref{sec:extrap_proxy}, we use six challenging reasoning benchmarks: \textbf{AIME} 2025 \citep{aime25} consisting of $30$ problems, \textbf{HMMT} Feb-25, Nov-25, and Feb-26 \citep{balunovic2025matharena} consisting a total of $93$ problems, \textbf{GPQA} main set \citep{rein2023gpqa} consisting a total of $448$ problems, \textbf{USACO} competitive programming \citep{shi2024usaco} consisting of $307$ problems, the computer-science and engineering subsets of \textbf{MMLU-Pro} \citep{wang2024mmlupro} consisting a total of $1379$ problems, and the hard non-science/engineering subset of \textbf{SuperGPQA} \citep{supergpqa} consisting of $382$ problems. 
Together these cover competition math, graduate-level science, broad professional knowledge, and competitive code.

In our experiments in \cref{sec:ranking_datasets} and \cref{sec:extrap_downstream}, we use OLMES which consists of MMLU \citep{hendryckstest2021}, HellaSwag \citep{zellers-etal-2019-hellaswag}, ARC Challenge \citep{Clark2018Think}, ARC Easy \citep{Clark2018Think}, PIQA \citep{Bisk_Zellers_Le_bras_Gao_Choi_2020}, CommonsenseQA \citep{talmor-etal-2019-commonsenseqa}, SocialIQA \citep{sap-etal-2019-social}, OpenBookQA \citep{mihaylov-etal-2018-suit}, BoolQ \citep{clark-etal-2019-boolq}, and WinoGrande \citep{Sakaguchi_Le_Bras_Bhagavatula_Choi_2020}.

\subsection{Other Experimental Details}\label{app:other_details}

The experiments in \Cref{sec:ranking_datasets} use DataDecide evaluation results with checkpoints of the default seed. The rBridge \citep{koh2025rbridge} results are based on our reproduction of their approach as described in their paper. While \citet{koh2025rbridge} experimented with GPT-4o \citep{openai2023gpt4} as the teacher, we instead work with an open-weights expert (Llama-3.3-70B) for more reliable access to logprobs. The results for the downstream task performance baseline are taken directly from the DataDecide huggingface repository.\footnote{\href{https://huggingface.co/datasets/allenai/DataDecide-eval-results}{https://huggingface.co/datasets/allenai/DataDecide-eval-results}}

The experiments in \Cref{sec:extrap_proxy} use MiniMax-M2.5 and Qwen3Next-80B as the experts. For pretraining extrapolation, we set the inner-split-fraction as $0.5$, fit on checkpoints up to step $350000$ and test the extrapolation at step $1300000$. To calculate the proxy metrics, since these are pretraining checkpoints, we use a standard chain-of-thought prompt which appends ``Let's think how to answer this question step by step.'' to the problem before the expert trajectory. For post-training extrapolation, we fit on checkpoints up to step $22000$ and test the extrapolation at step $40000$.

\subsection{Software and Compute Requirements}\label{app:compute}

Our code is implemented in PyTorch \citep{pytorch} and makes use of the HuggingFace Transformers library \citep{huggingface} and the vLLM library \citep{kwon2023efficient} for running efficient inference locally on LLMs. All benchmark evaluation and proxy metric calculation experiments were done on our cluster with $4$ NVIDIA H100 GPUs with $80$ GB memory. Experiments with Kimi-K2.5 were carried out using Together API.\footnote{\href{https://api.together.xyz/}{https://api.together.xyz/}}. Obtaining model generations for benchmark evaluations takes around $2$ hours on average for a single seed on a benchmark of $100$ problems. Computing proxy metrics is extremely fast and takes less than $5$ minutes per evaluation.

\section{Additional Results and Discussion}\label{app:additional_results}

\subsection{Cross-Family Model Selection}
\label{app:ranking_llms}

This section provides additional analysis of the cross-family model selection experiment (\S\ref{sec:ranking_llms}). We examine which proxy metrics are selected under cross-validation and oracle access, how the ranking signal compares to loss-based baselines, and how robust the results are to the number of held-out tasks and the fraction of models available for selection.

\begin{table}[t]
\centering
\caption{\textbf{Oracle proxy selection (upper bound).} Spearman $\rho$ when the proxy is selected using downstream scores from all six benchmarks and the full model population. Because the proxy is fit and evaluated on the same tasks, these numbers upper-bound the cross-validated results in \cref{tab:main_results}.}
\label{tab:oracle_results}
\setlength{\tabcolsep}{5pt}
\renewcommand{\arraystretch}{1.15}
\begin{tabular}{l c cccccc}
\toprule
\textbf{Proxy} & \textbf{Mean $\rho$} & AIME & GPQA & HMMT & MMLU-Pro & SuperGPQA & USACO \\
\midrule
Univariate proxy & 0.62 & 0.51 & 0.64 & 0.43 & 0.73 & 0.81 & 0.62 \\
$3$-sparse proxy & 0.88 & 0.89 & 0.97 & 0.89 & 0.88 & 0.89 & 0.78 \\
\bottomrule
\end{tabular}
\end{table}

\begin{table}[t]
\centering
\caption{\textbf{Proxy metrics selected under oracle access.} The specific proxy metrics and coefficients chosen when fitting on all six benchmarks and the full model population simultaneously. For the univariate proxy, the selected metric is inverse-frequency-weighted top-1 accuracy, the same metric that dominates the cross-validated selection (\cref{fig:feature_heatmap}). For the 3-sparse proxy, the negative coefficient on frequency-weighted top-5 accuracy penalizes models whose accuracy concentrates on frequent (and therefore easy) tokens, complementing the two positive components that reward uncertainty-aware agreement with the expert.}
\label{tab:oracle_proxies}
\setlength{\tabcolsep}{6pt}
\renewcommand{\arraystretch}{1.15}
\begin{tabular}{l l l}
\toprule
\textbf{Model} & \textbf{Selected proxy metric} & \textbf{Coefficient} \\
\midrule
Univariate & \texttt{inverse\_frequency\,/\,top\_1\_accuracy} & $+1$ \\
\midrule
\multirow{3}{*}{$3$-sparse} & \texttt{expert\_disagreement\,/\,entropy} & $+1$ \\
 & \texttt{inverse\_entropy\,/\,margin} & $+1$ \\
 & \texttt{frequency\,/\,top\_5\_accuracy} & $-1$ \\
\bottomrule
\end{tabular}
\end{table}

\begin{figure}[t]
\centering
\begin{subfigure}[t]{0.49\textwidth}
    \includegraphics[width=\textwidth]{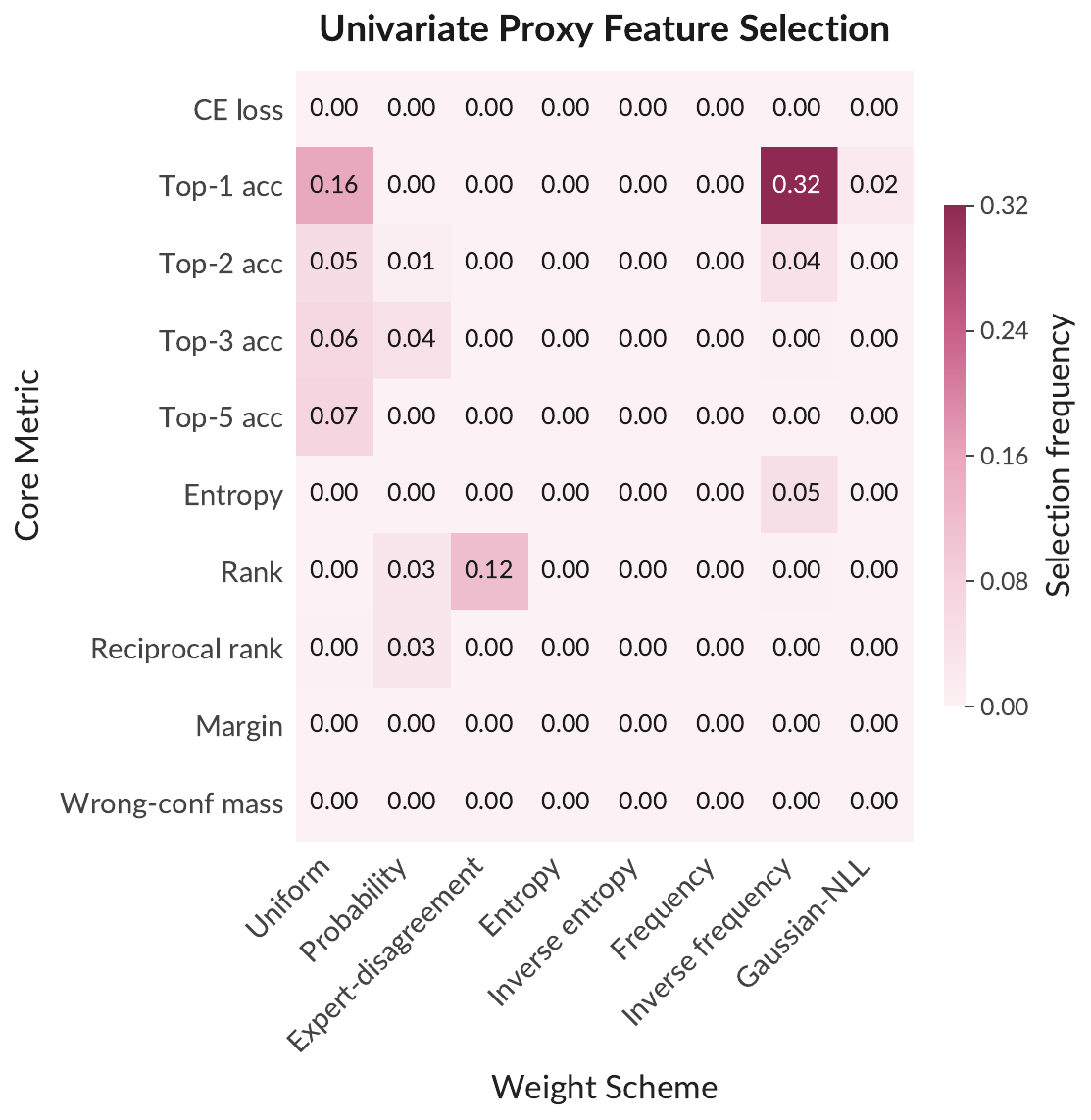}
\end{subfigure}
\hfill
\begin{subfigure}[t]{0.49\textwidth}
    \includegraphics[width=\textwidth]{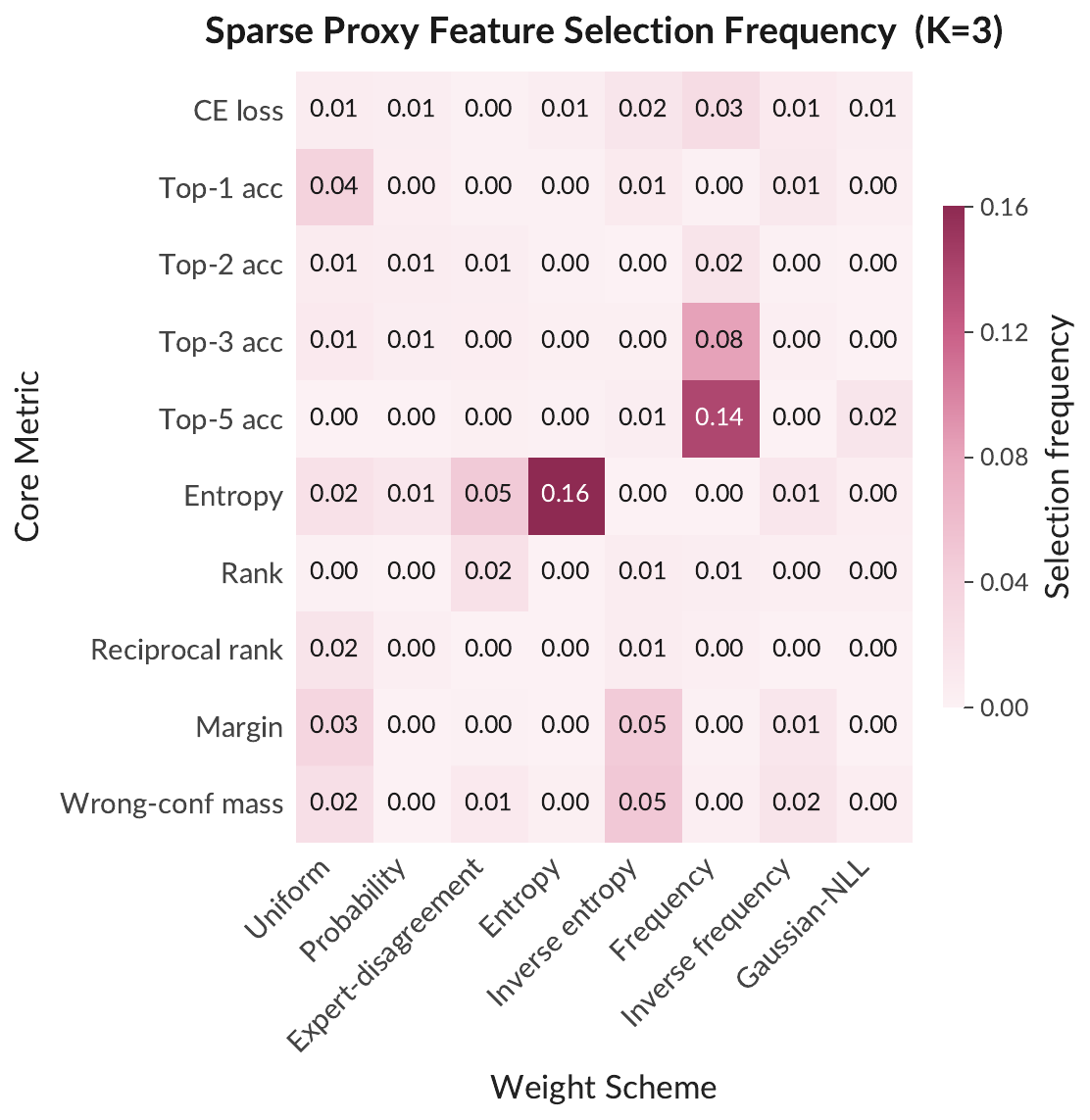}
\end{subfigure}
\caption{Proxy metric selection frequency (normalized) for univariate (\emph{left}) and 3-sparse proxy (\emph{right}). Darker cells are chosen more often. In both cases, a small cluster of proxy metrics accounts for almost all selections. The bulk of the 80 proxy metrics are unused.}
\label{fig:feature_heatmap}
\end{figure}

\paragraph{Which proxy metrics carry the ranking signal?}
\Cref{fig:feature_heatmap} visualizes the frequency with which each of the 80 proxy metrics is selected across all leave-2-tasks-out folds and model-subsampling seeds. In both the univariate and 3-sparse settings, the selection mass concentrates on a small number of cells, with the majority of the 80 proxy metrics never selected.

For the univariate proxy (\cref{fig:feature_heatmap}, left), inverse-frequency-weighted top-1 accuracy accounts for $32\%$ of all selections, more than twice the frequency of any other cell. This metric measures whether the candidate model's top prediction matches the expert token, but only at positions where the expert token is rare. Frequent tokens, such as punctuation, articles, and common function words, are predicted well by nearly all models and therefore carry little discriminative signal. Upweighting rare tokens isolates the positions where models are most likely to differ: variable names in code, technical terms in science, or key numerical quantities in mathematics. The remaining selection mass spreads across other top-$k$ accuracy variants (uniform top-1 at $0.16$, uniform top-5 at $0.07$) and expert-disagreement-weighted rank ($0.12$), all of which measure the same broad phenomenon, whether the model agrees with the expert, through slightly different lenses.

\begin{figure}[t]
\centering
\begin{subfigure}[t]{0.48\textwidth}
    \includegraphics[width=\textwidth]{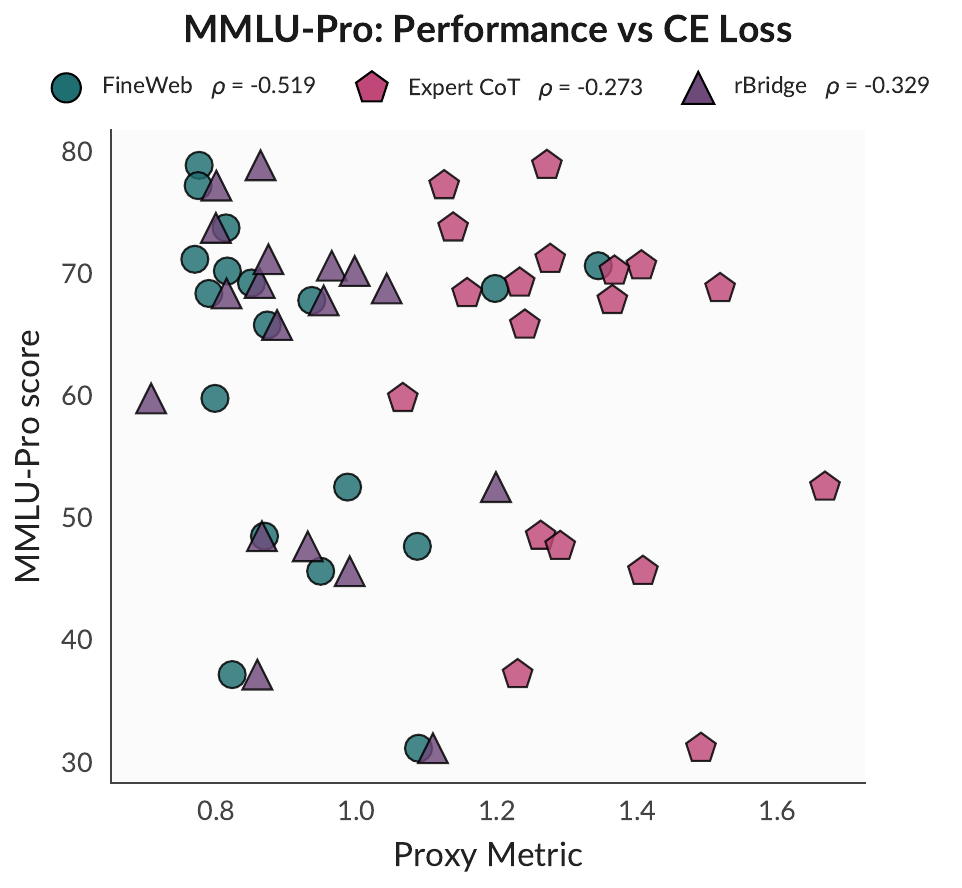}
\end{subfigure}
\hfill
\begin{subfigure}[t]{0.51\textwidth}
    \includegraphics[width=\textwidth]{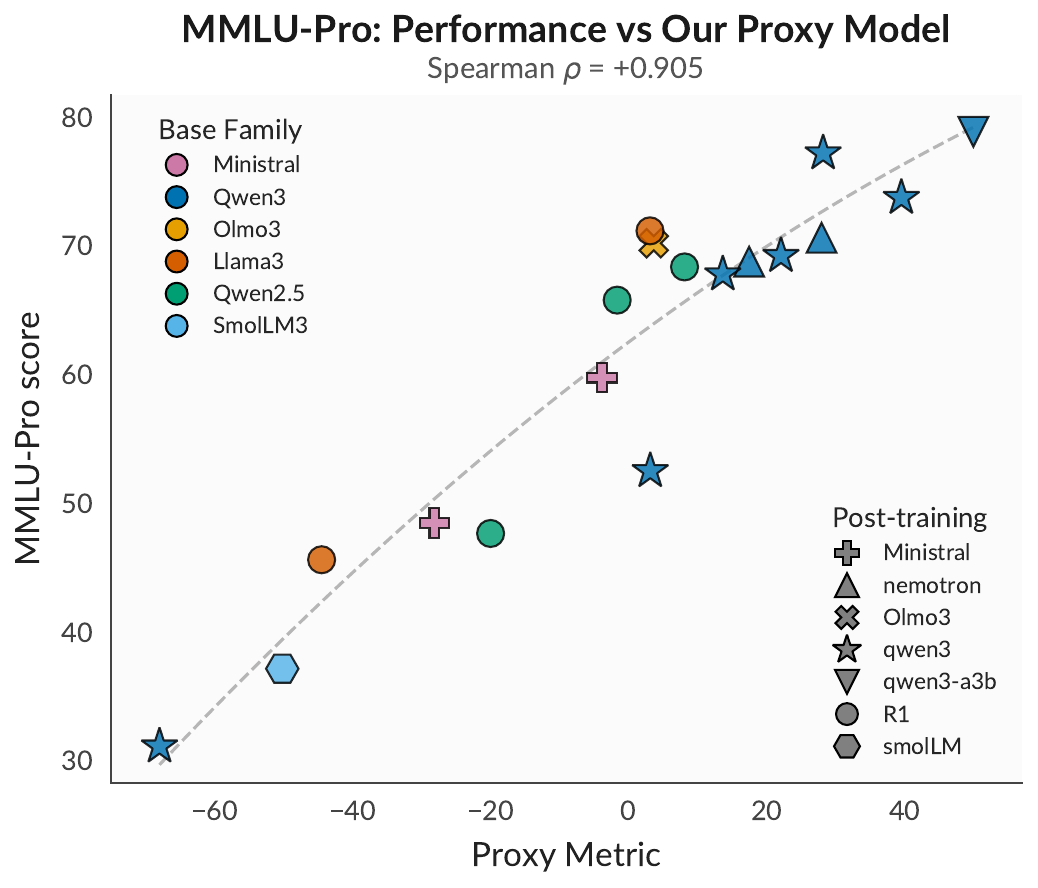}
\end{subfigure}
\caption{\textbf{Likelihood-based baselines against a learned proxy.} \emph{Left:} the three loss-based baselines (FineWeb cross-entropy loss, uniform expert-trajectory cross-entropy loss, and rBridge) plotted against MMLU-Pro accuracy across 18 language models. Low loss is a weak and non-monotonic indicator of downstream ranking across model families and post-training recipes. \emph{Right:} the learned RankSVM (linear) proxy, evaluated on MMLU-Pro in a fold where MMLU-Pro was held out, produces a nearly monotonic relationship with downstream accuracy. The same proxy was not exposed to MMLU-Pro during fitting.}
\label{fig:unseen_scatter}
\end{figure}

For the 3-sparse proxy (\cref{fig:feature_heatmap}, right), the pattern shifts. The two most frequently selected cells are entropy-weighted entropy ($0.16$) and frequency-weighted top-5 accuracy ($0.14$). Entropy-weighted entropy upweights positions where the candidate model is uncertain and asks how diffuse its distribution is at those positions. A model that is uncertain at the right places, the critical reasoning steps, and concentrated elsewhere, is one that has learned the structure of the task without necessarily being able to solve it. Frequency-weighted top-5 accuracy, by contrast, measures agreement with the expert on common tokens. These two signals are complementary: the first captures behavior at hard, high-entropy positions, while the second captures baseline competence on the easy positions. The oracle 3-sparse proxy (\cref{tab:oracle_proxies}) confirms this complementarity by including both an uncertainty-aware component (expert-disagreement-weighted entropy, coefficient $+1$) and a negative coefficient on frequency-weighted top-5 accuracy ($-1$), which penalizes models whose agreement with the expert concentrates on frequent tokens where all models perform well.

\begin{figure}[t]
\centering
\includegraphics[width=0.9\textwidth]{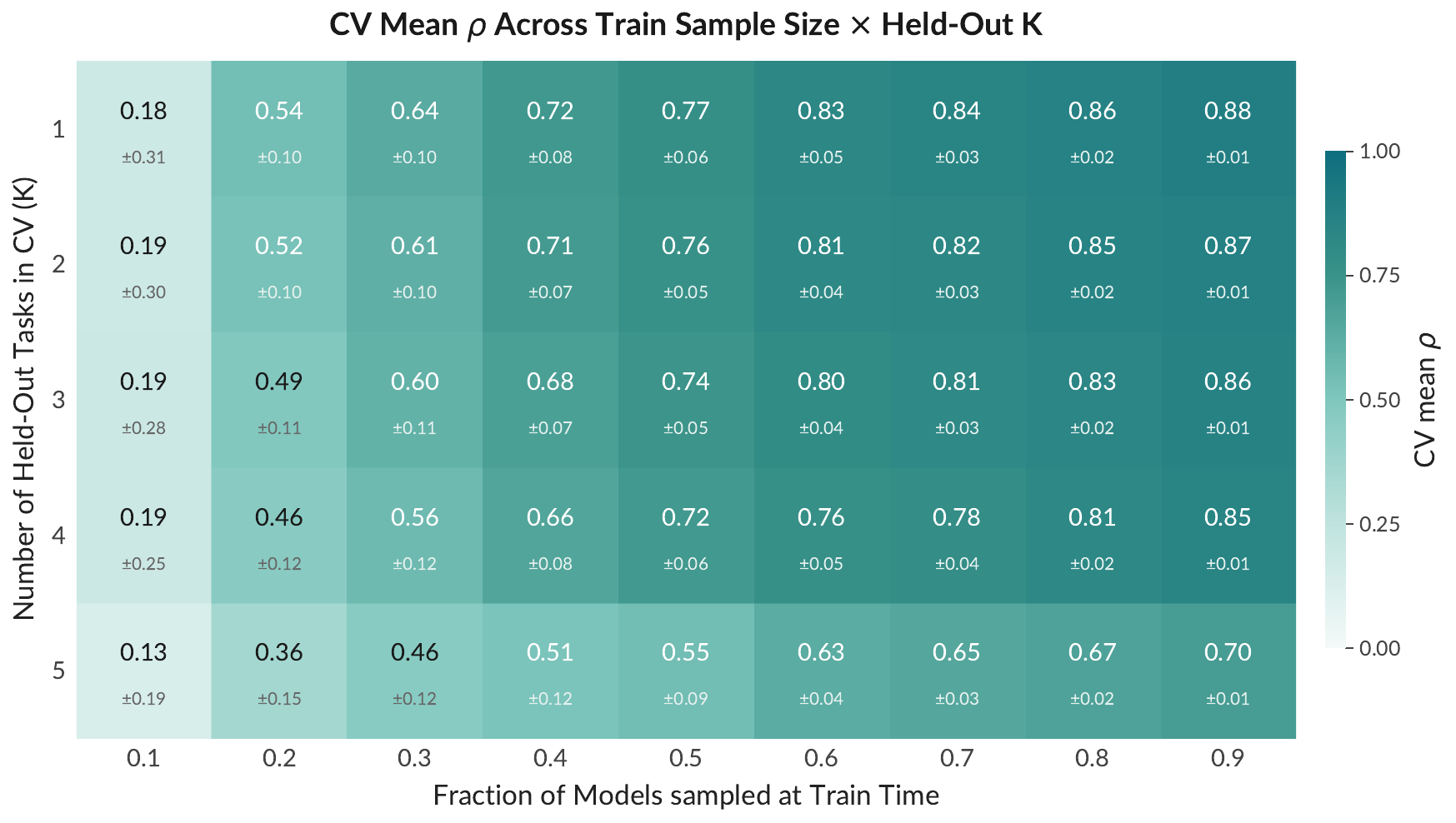}
\caption{Performance of the linear RankSVM proxy as we vary the number of held-out tasks and the fraction of models used for selection. Each cell reports the mean Spearman $\rho$ on held-out tasks.}
\label{fig:sweep_held_perc_linear}
\end{figure}

\paragraph{Oracle upper bounds.}
\Cref{tab:oracle_results} reports the Spearman $\rho$ when the proxy is selected using downstream scores from all six benchmarks and the full model population. Because the proxy is fit and evaluated on the same data, these numbers upper-bound the cross-validated results in \cref{tab:main_results}. The univariate oracle reaches $\rho = 0.62$, with substantial variation across tasks (HMMT at $0.43$, SuperGPQA at $0.81$). The 3-sparse oracle reaches $\rho = 0.88$ with per-task correlations between $0.78$ and $0.97$, indicating that three complementary proxy metrics contain nearly all the ranking information present in the 80-dimensional library. The gap between the oracle ($0.88$) and the cross-validated 3-sparse result ($0.78$) reflects the difficulty of selecting the right triplet without access to the target task, and suggests that better proxy selection procedures could close part of this gap.

\Cref{tab:oracle_proxies} lists the specific proxy metrics selected under oracle access. The univariate oracle selects inverse-frequency-weighted top-1 accuracy, the same metric that dominates the cross-validated heatmap, providing evidence that the cross-validated selection procedure converges to a genuinely informative signal rather than overfitting to the held-in tasks. The 3-sparse oracle selects three metrics from three distinct regimes of model--expert alignment: expert-disagreement-weighted entropy (an uncertainty signal at positions where the model disagrees with the expert), inverse-entropy-weighted margin (a confidence signal at positions where the model is peaked), and frequency-weighted top-5 accuracy with a negative sign (a correction that discounts agreement on easy tokens). This combination is consistent with the view that ranking heterogeneous models requires measuring not just how often a model agrees with the expert, but how its uncertainty and confidence are distributed across token positions of varying difficulty.

\paragraph{Loss-based baselines fail across model families.}
\Cref{fig:unseen_scatter} contrasts the loss-based baselines with the learned proxy on MMLU-Pro. The left panel plots three loss-based signals, FineWeb cross-entropy ($\rho = -0.52$), uniform expert-trajectory cross-entropy ($\rho = -0.27$), and rBridge ($\rho = -0.33$), against MMLU-Pro accuracy. None of these produces a coherent ranking. Models with very different downstream scores overlap at similar loss values, and the overall relationship is weak and non-monotonic. The right panel shows the linear RankSVM proxy evaluated on a fold where MMLU-Pro was held out. Despite never seeing MMLU-Pro scores during fitting, the proxy produces a nearly monotonic relationship with downstream accuracy (Spearman $\rho = 0.91$), and this relationship holds across all six base families and seven post-training recipes. The contrast illustrates the central claim of \S\ref{sec:ranking_llms}: a scalar summary of fit to generic text is a poor predictor of relative downstream performance across a heterogeneous model population, but a small number of task-conditioned token-level statistics can recover the ranking reliably.

\begin{figure}[t]
\centering
\includegraphics[width=0.9\textwidth]{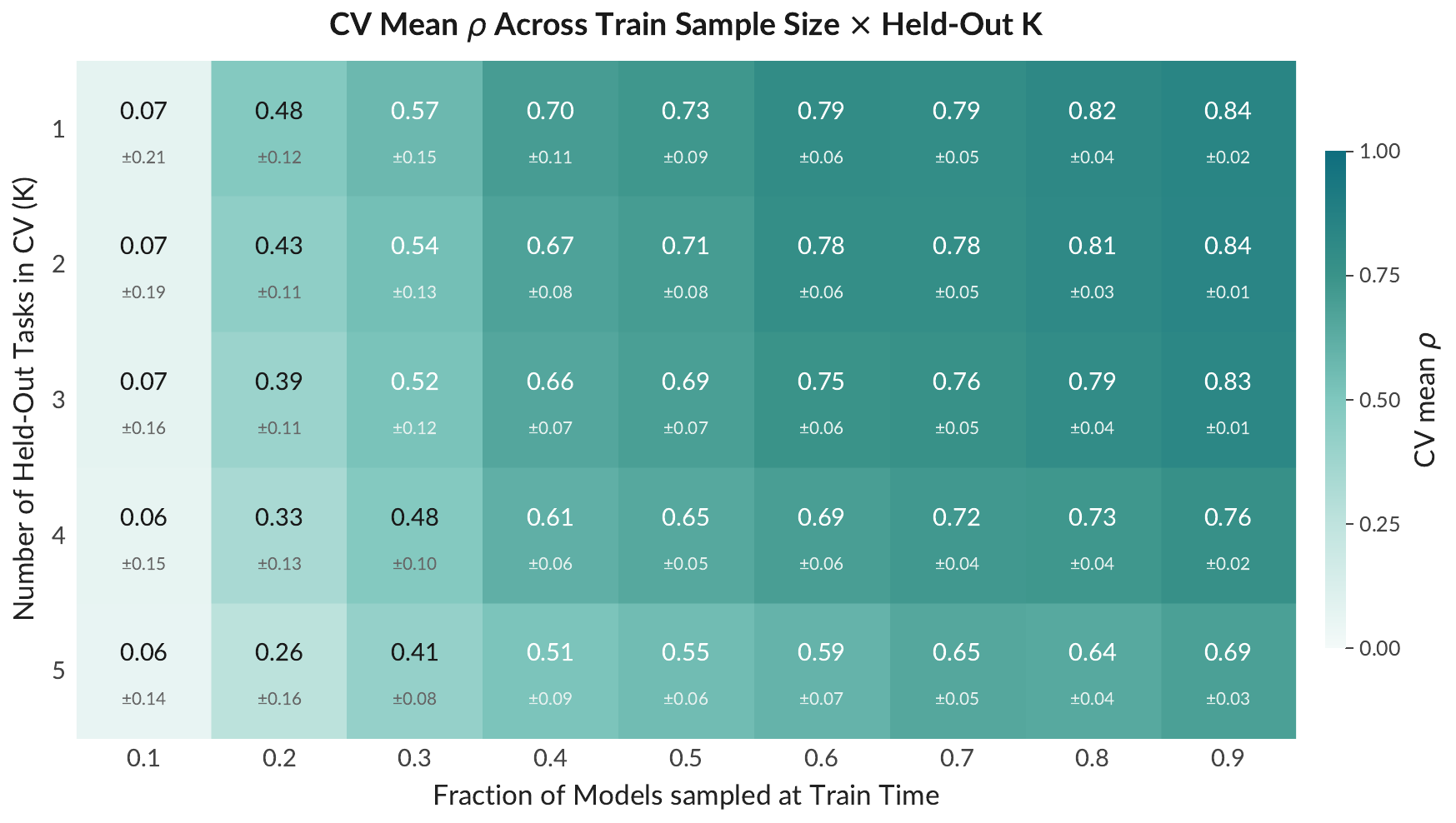}
\caption{Performance of the 3-sparse proxy as we vary the number of held-out tasks and the fraction of models used for selection. Each cell reports the mean Spearman $\rho$ on held-out tasks.}
\label{fig:sweep_held_perc_sparse}
\end{figure}

\begin{figure}[t]
\centering
\includegraphics[width=0.6\textwidth]{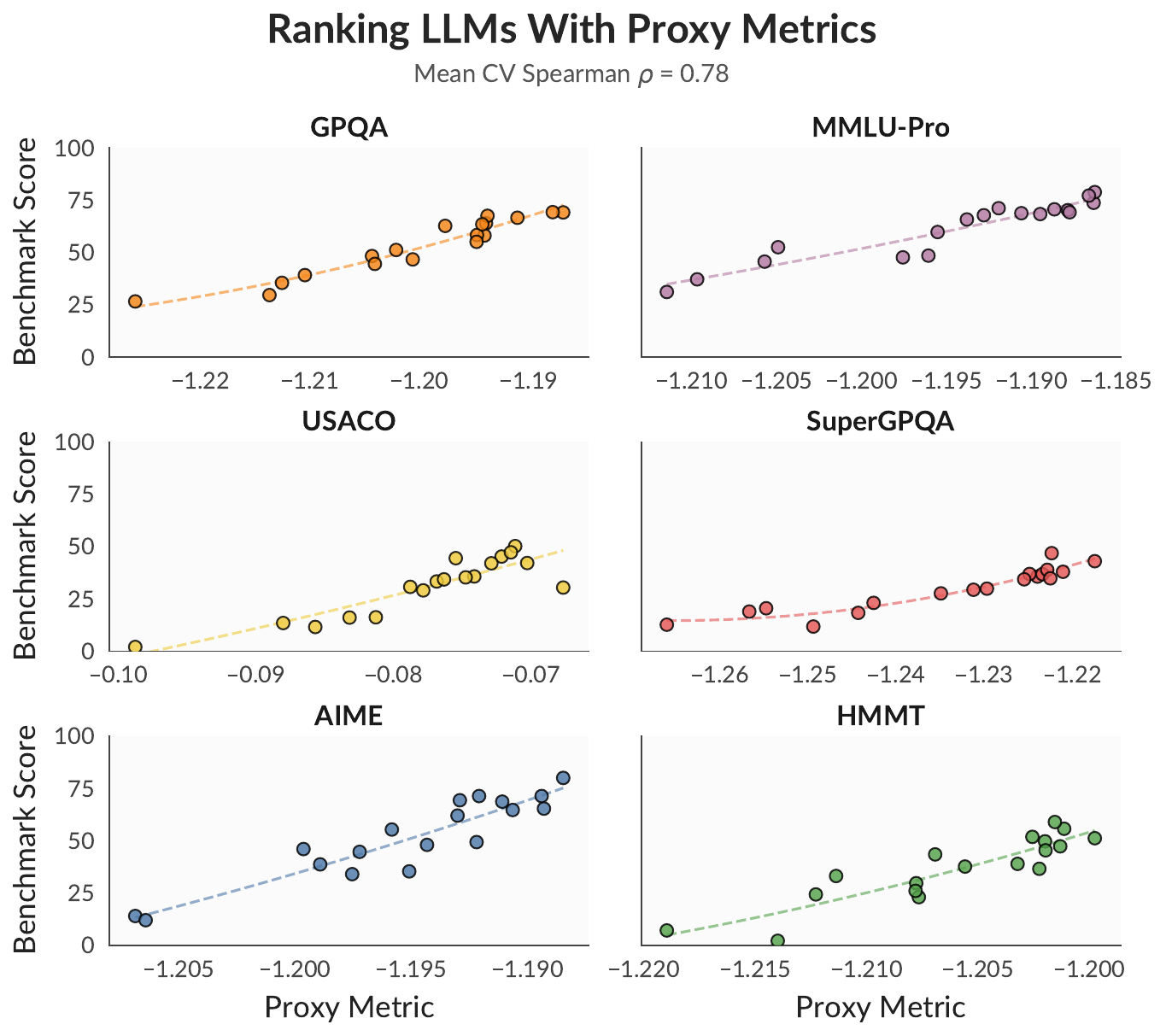}
\caption{\textbf{Ranking LLMs with the $3$-sparse proxy.} Downstream accuracy vs.\ proxy score for each of the six benchmarks on a randomly sampled held-out fold. Same format as \cref{fig:main} (left) but using the $3$-sparse proxy instead of the linear RankSVM.}
\label{fig:3_sparse_scatter_all}
\end{figure}

\paragraph{Robustness to held-out tasks and model fraction.}
\Cref{fig:sweep_held_perc_linear,fig:sweep_held_perc_sparse} show how the linear RankSVM and 3-sparse proxies degrade as we increase the number of held-out tasks (rows) and decrease the fraction of models available for selection (columns). Performance degrades gracefully in both dimensions. For the linear RankSVM, holding out $K = 2$ tasks with $60\%$ of models yields $\rho = 0.81$, and even holding out $K = 3$ tasks with $50\%$ of models still achieves $\rho = 0.74$. Variance decreases steadily with more models, from $\pm 0.10$ at $20\%$ to $\pm 0.01$ at $90\%$. The 3-sparse proxy shows a similar pattern. The one regime where performance collapses is $K = 5$ held-out tasks (i.e., selecting from a single held-in task), which drops to $\rho \approx 0.36$--$0.70$ depending on model fraction. This is expected: with only one task for selection, the procedure cannot distinguish proxy metrics that generalize from those that happen to work on that particular task.

\paragraph{Visual comparison of proxy models.}
\Cref{fig:3_sparse_scatter_all} plots downstream accuracy against the 3-sparse proxy score across all six benchmarks in a randomly sampled held-out fold, providing a counterpart to \cref{fig:main} (left). The 3-sparse proxy produces monotonic trends on all six tasks, with somewhat tighter scatter than the linear RankSVM on GPQA and MMLU-Pro but slightly more variance on HMMT. This is consistent with the quantitative comparison in \cref{tab:main_results}, where the two models achieve similar mean $\rho$ ($0.78$ vs.\ $0.81$) but differ on individual tasks.

\begin{figure}[t]
\centering
\begin{subfigure}[t]{0.49\textwidth}
    \includegraphics[width=\textwidth]{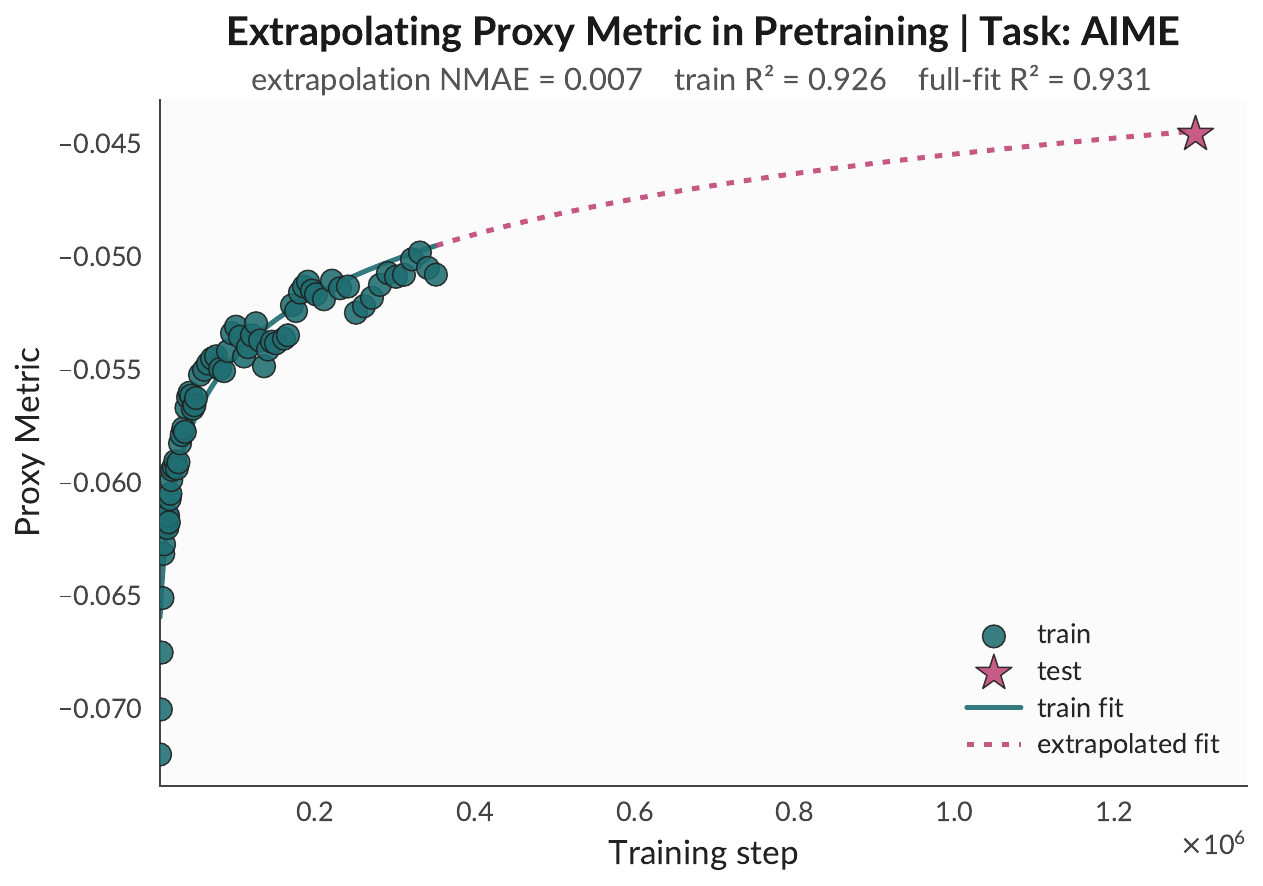}
\end{subfigure}
\hfill
\begin{subfigure}[t]{0.49\textwidth}
    \includegraphics[width=\textwidth]{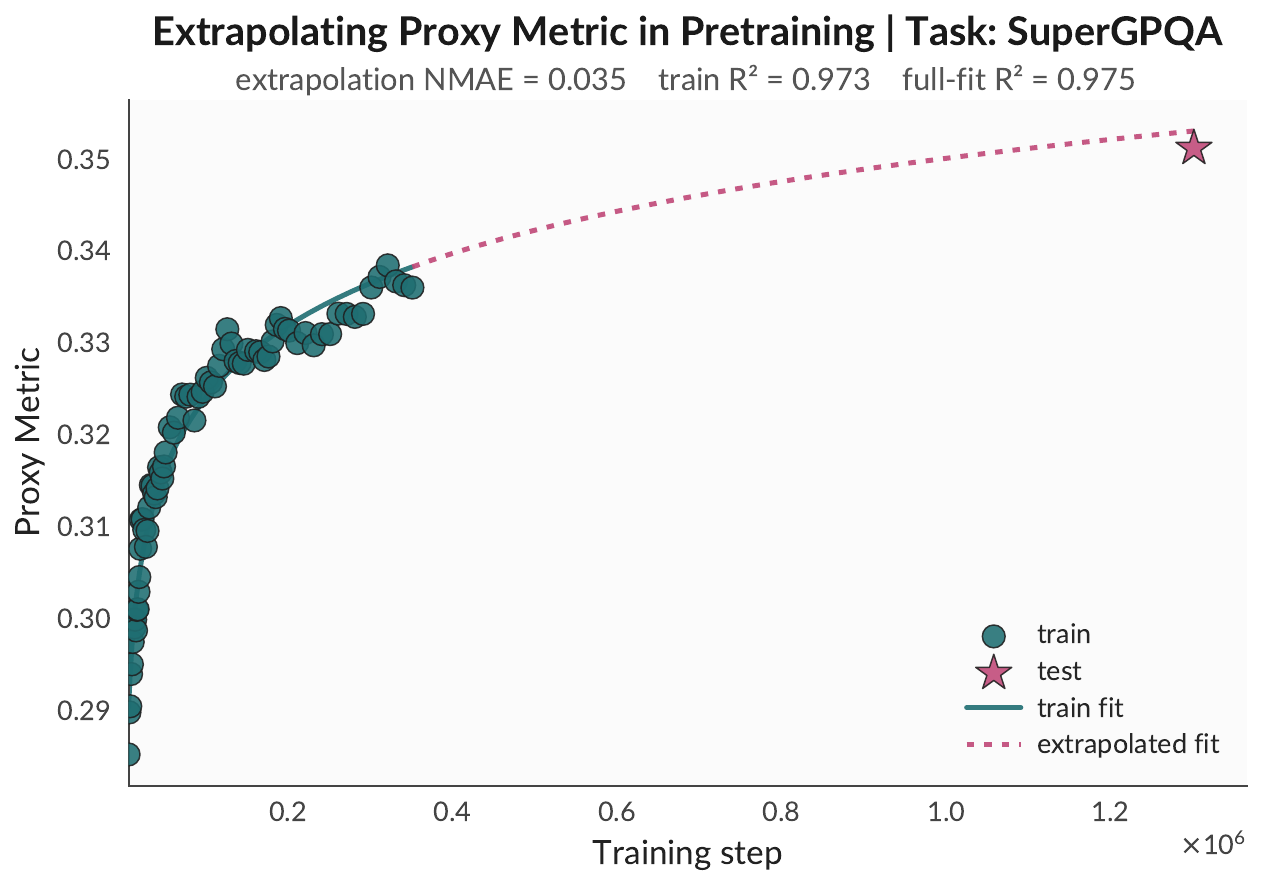}
\end{subfigure}
\caption{\textbf{Pretraining extrapolation on AIME and SuperGPQA.} Same protocol as \cref{fig:extrapolation} (left) for the two benchmarks not shown in the main figure. The power-law fits from the training window (filled markers) extrapolate to the held-out checkpoint (star) on both tasks.}
\label{fig:extrapolation_pre_aime_supergpqa}
\end{figure}

\begin{figure}[t]
\centering
\includegraphics[width=0.5\textwidth]{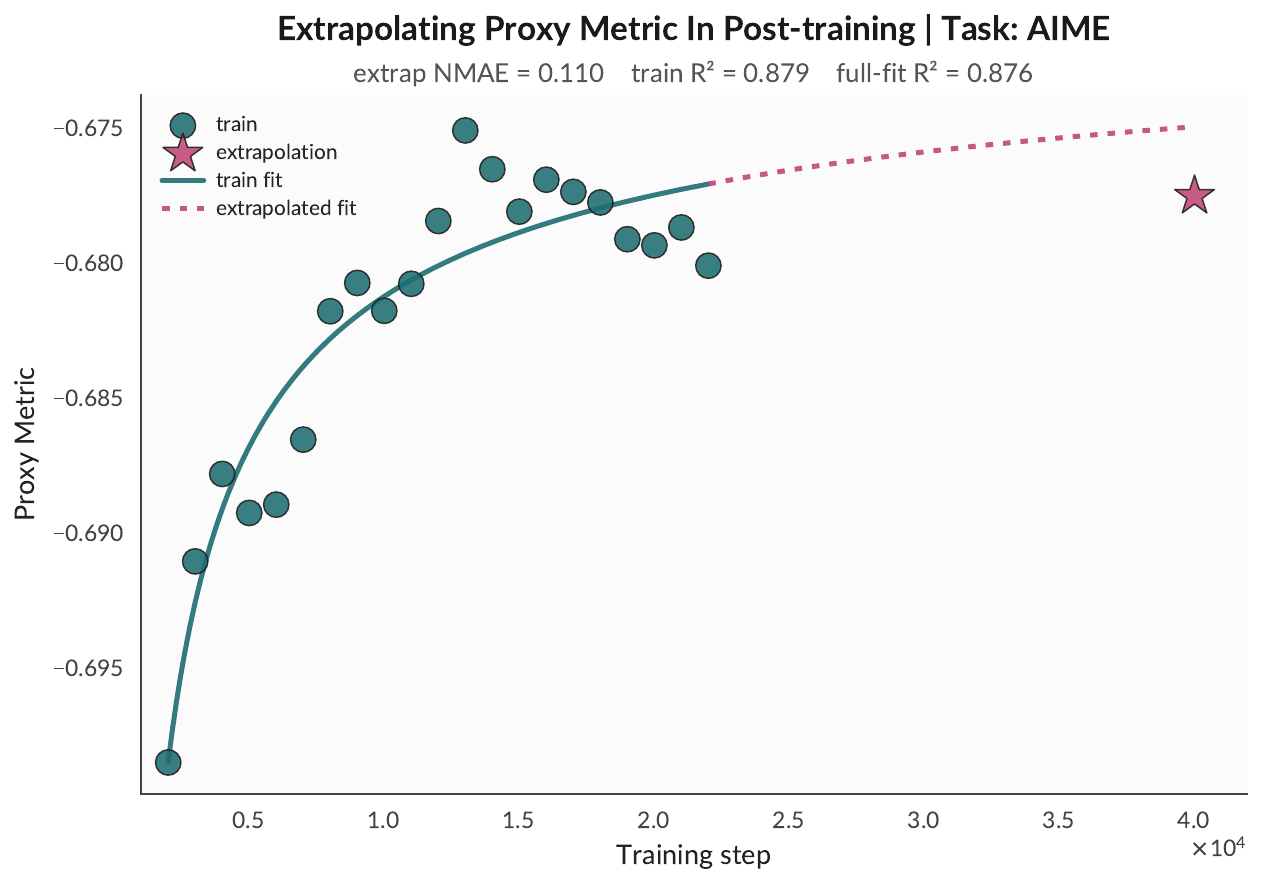}
\caption{\textbf{Post-training extrapolation on AIME.} Same protocol as \cref{fig:extrapolation} (right). The proxy metric evolves monotonically with post-training step and the power-law fit extrapolates to the held-out checkpoint.}
\label{fig:extrapolation_post_aime}
\end{figure}

\begin{figure}[t]
\centering
\begin{subfigure}[t]{0.49\textwidth}
    \includegraphics[width=\textwidth]{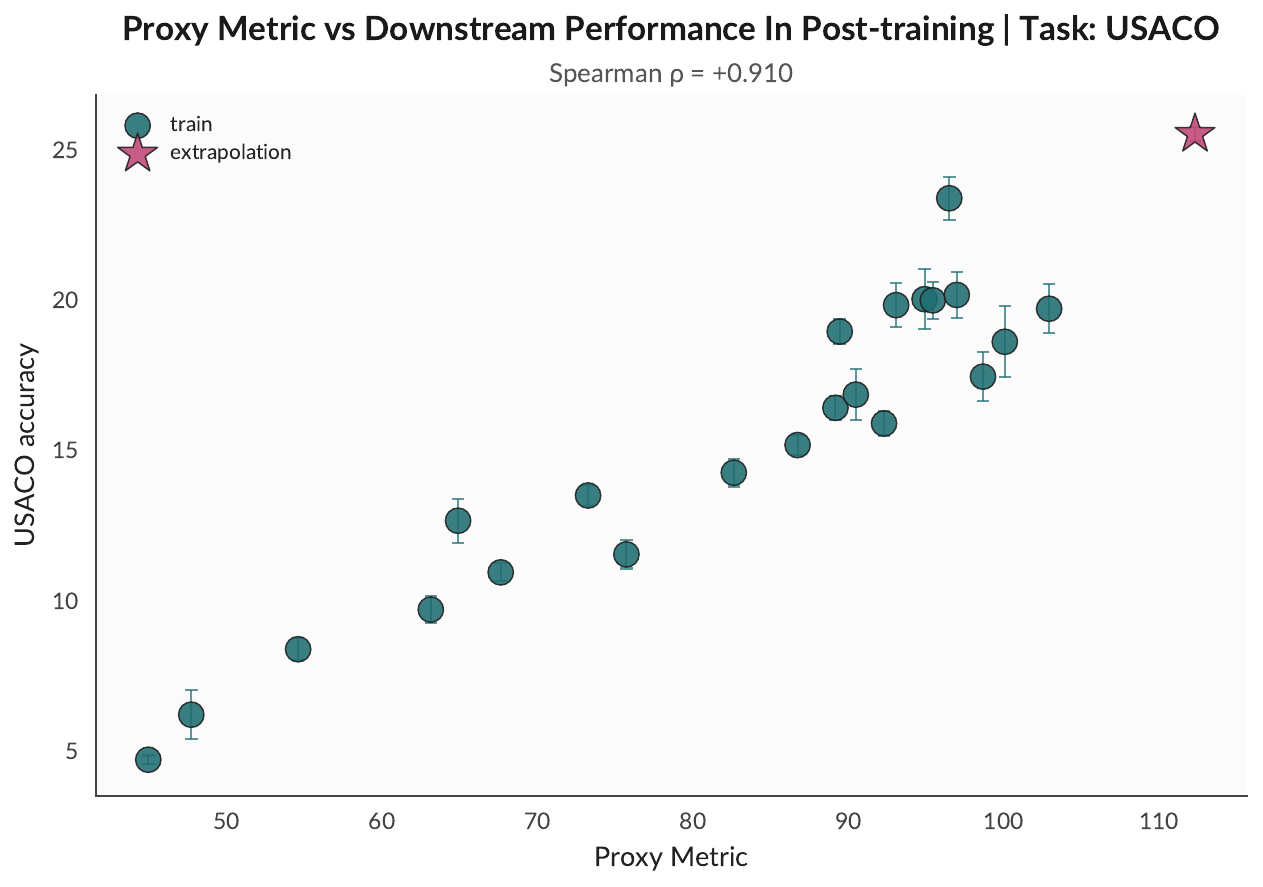}
\end{subfigure}
\hfill
\begin{subfigure}[t]{0.49\textwidth}
    \includegraphics[width=\textwidth]{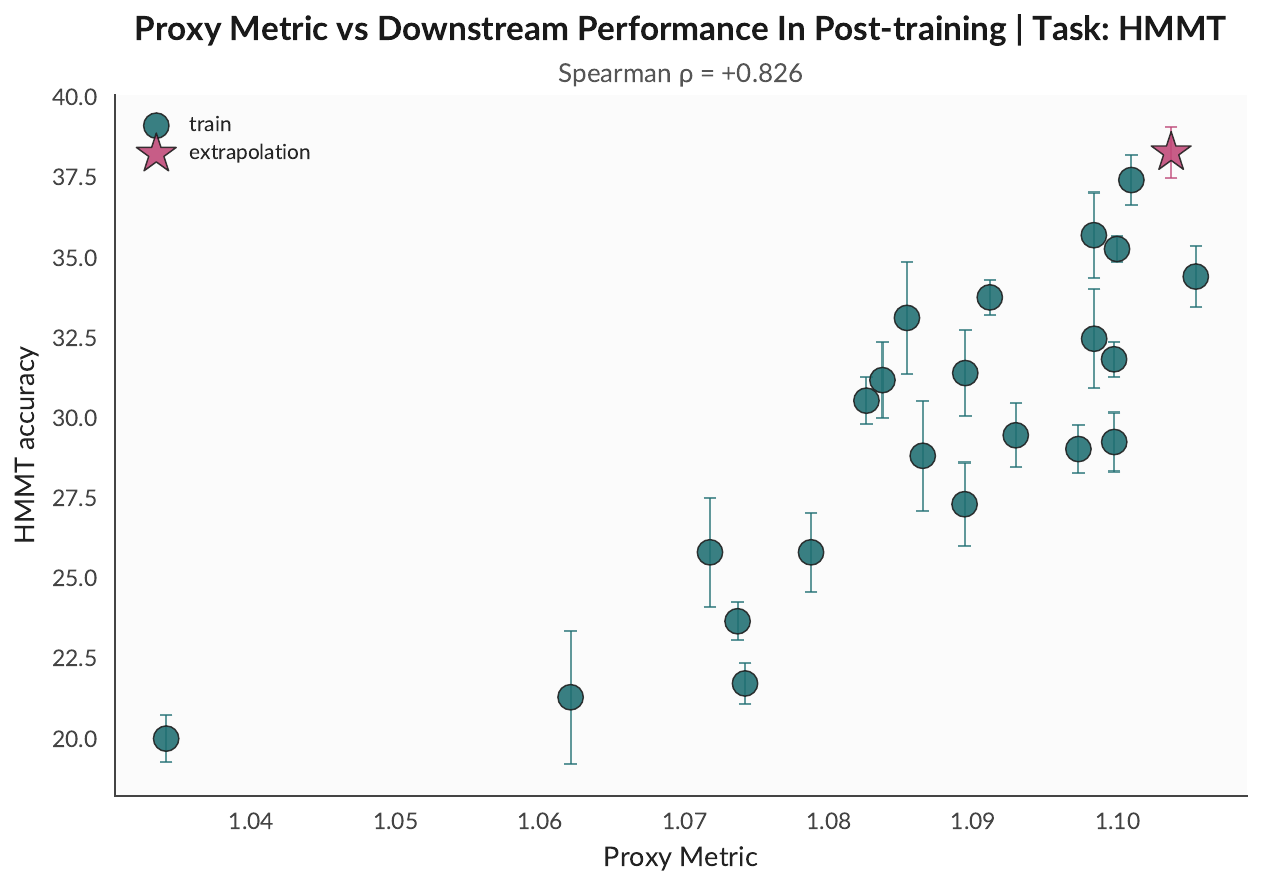}
\end{subfigure}
\caption{\textbf{Proxy metric vs.\ downstream accuracy at post-training checkpoints.} The best selected univariate proxy is plotted against downstream accuracy on USACO (left) and HMMT (right) across post-training checkpoints of OLMo-3-7B-Think. The strong monotonic relationship confirms that the extrapolated proxy tracks the ranking of interest.}
\label{fig:extrapolation_post_scatter_usaco_hmmt}
\end{figure}

\begin{figure}[t]
\centering
\includegraphics[width=0.7\textwidth]{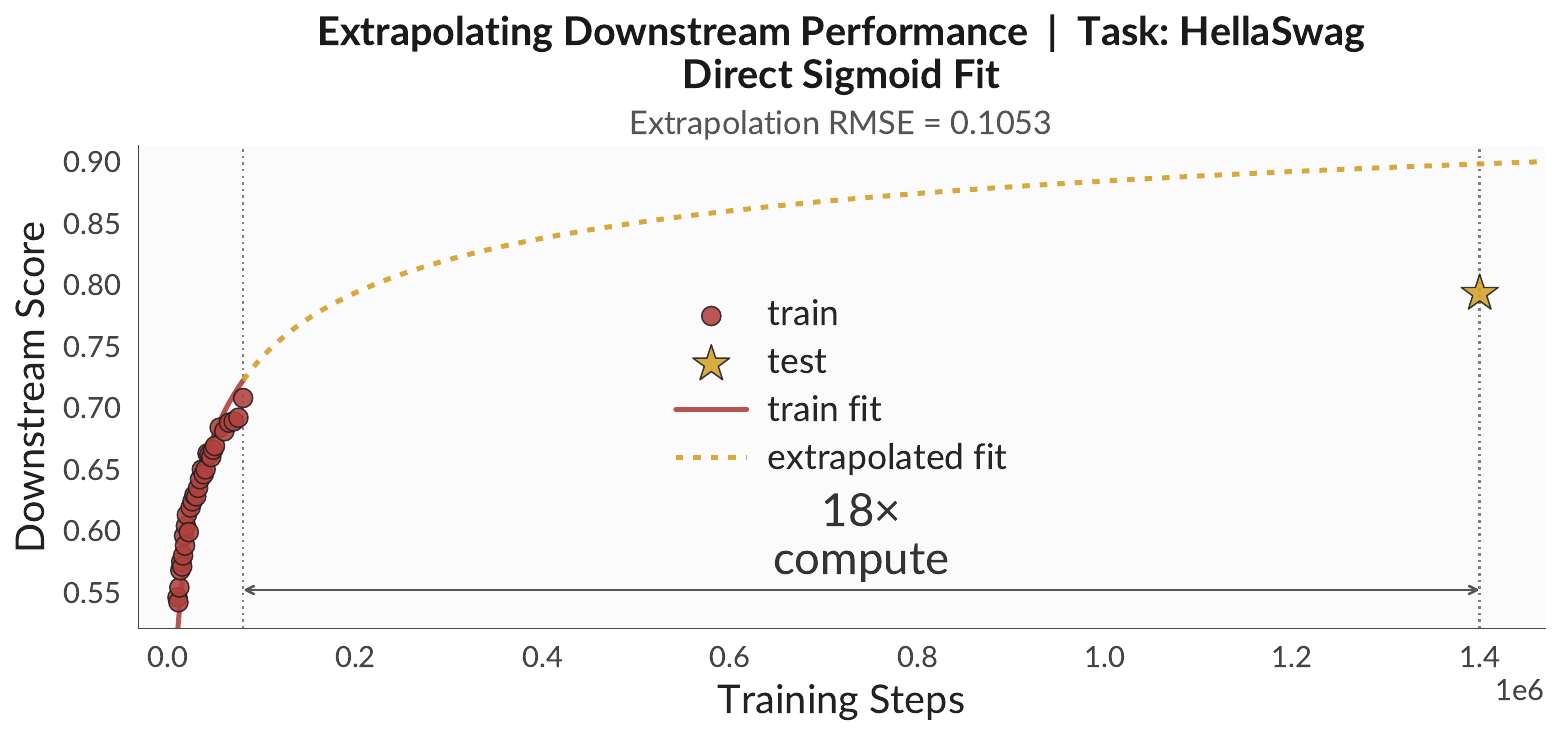}
\caption{\textbf{Direct sigmoid extrapolation of HellaSwag accuracy.} Accuracy is fit as a sigmoid of $\log_{10}(\text{steps})$ following \citet{owen2024predictablelanguagemodelbenchmark}. Circles are the training window (up to $80$K steps), the star is the held-out checkpoint at $1.4$M steps. The fit overshoots the held-out accuracy (RMSE $= 0.11$).}
\label{fig:sigmoid_hellaswag}
\end{figure}

\begin{figure}[t]
\centering
\begin{subfigure}[t]{0.325\textwidth}
    \includegraphics[width=\textwidth]{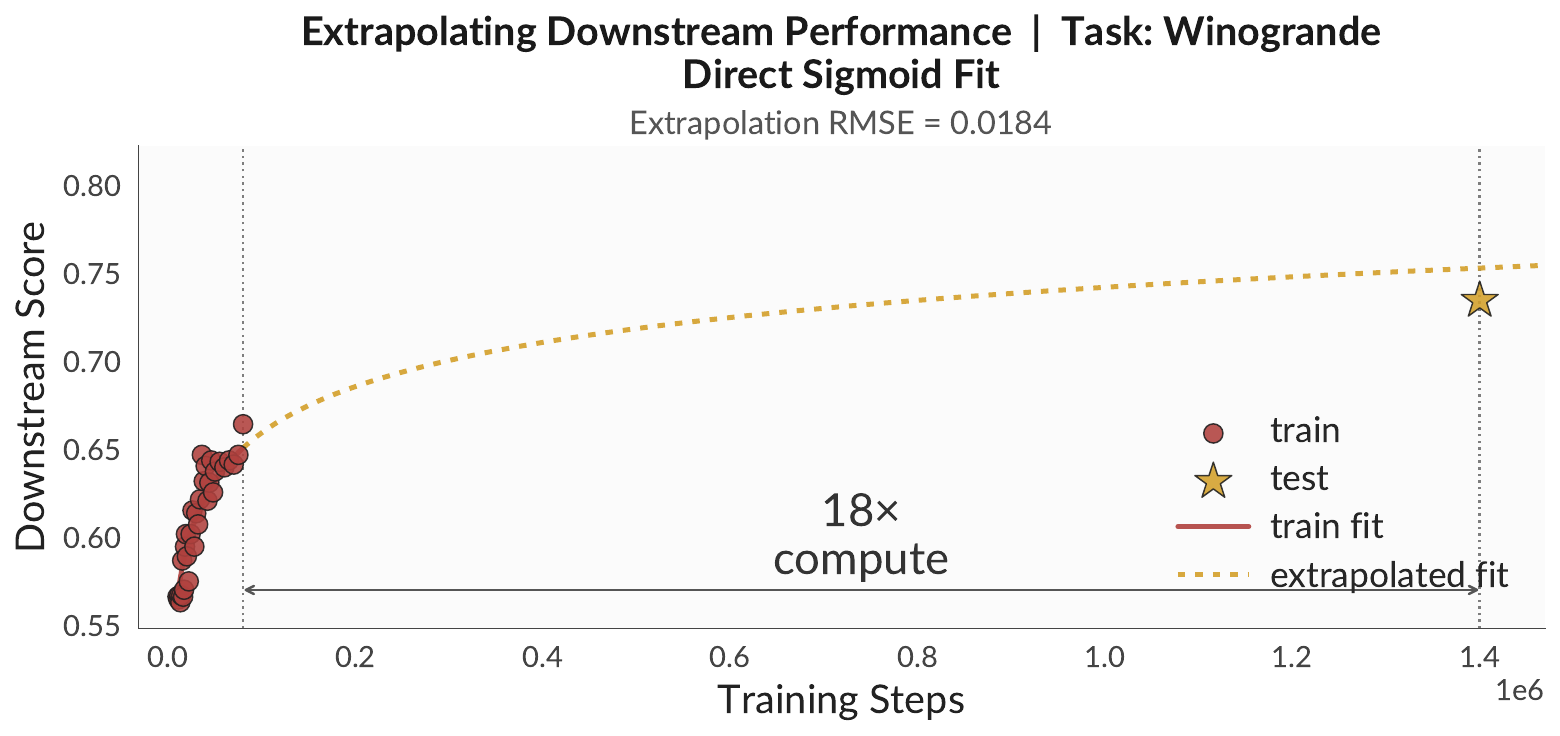}
\end{subfigure}
\hfill
\begin{subfigure}[t]{0.325\textwidth}
    \includegraphics[width=\textwidth]{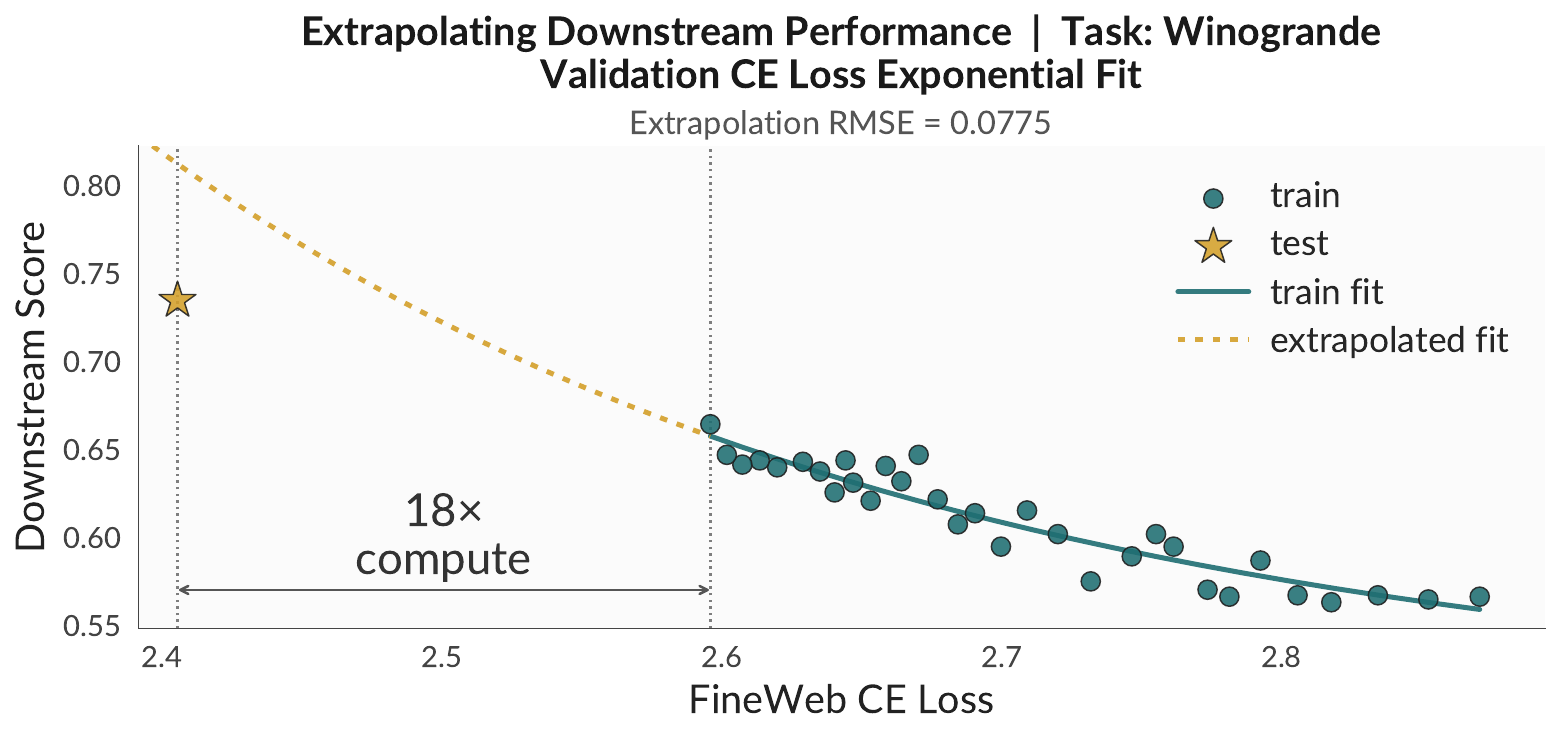}
\end{subfigure}
\hfill
\begin{subfigure}[t]{0.325\textwidth}
    \includegraphics[width=\textwidth]{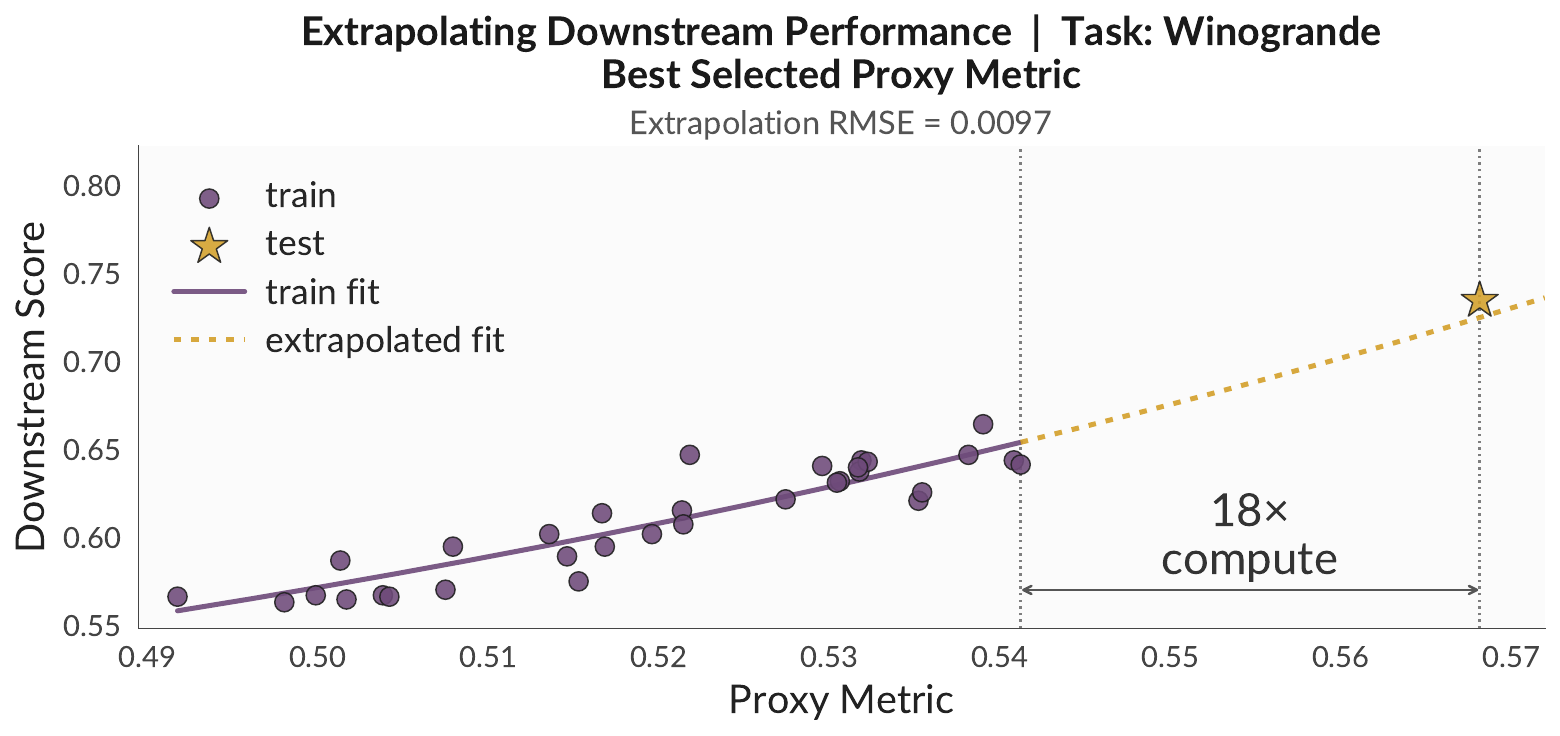}
\end{subfigure}
\caption{\textbf{Extrapolating Winogrande accuracy along the pretraining trajectory of OLMo-3-7B.} Circles are the training window (up to $80$K steps), the star is the held-out checkpoint at $1.4$M steps (${\sim}18\times$ the training compute). \emph{Left:} accuracy vs. $\log_{10}(\text{steps})$, fit with a sigmoid (RMSE $= 0.02$). \emph{Centre:} accuracy vs.\ CE loss on FineWeb, fit with an exponential (RMSE $= 0.08$). \emph{Right:} accuracy vs.\ the best univariate proxy, fit with a power law (RMSE $= 0.01$).}
\label{fig:proxy_vs_downstream_winogrande}
\end{figure}

\begin{figure}[t]
\centering
\begin{subfigure}[t]{0.325\textwidth}
    \includegraphics[width=\textwidth]{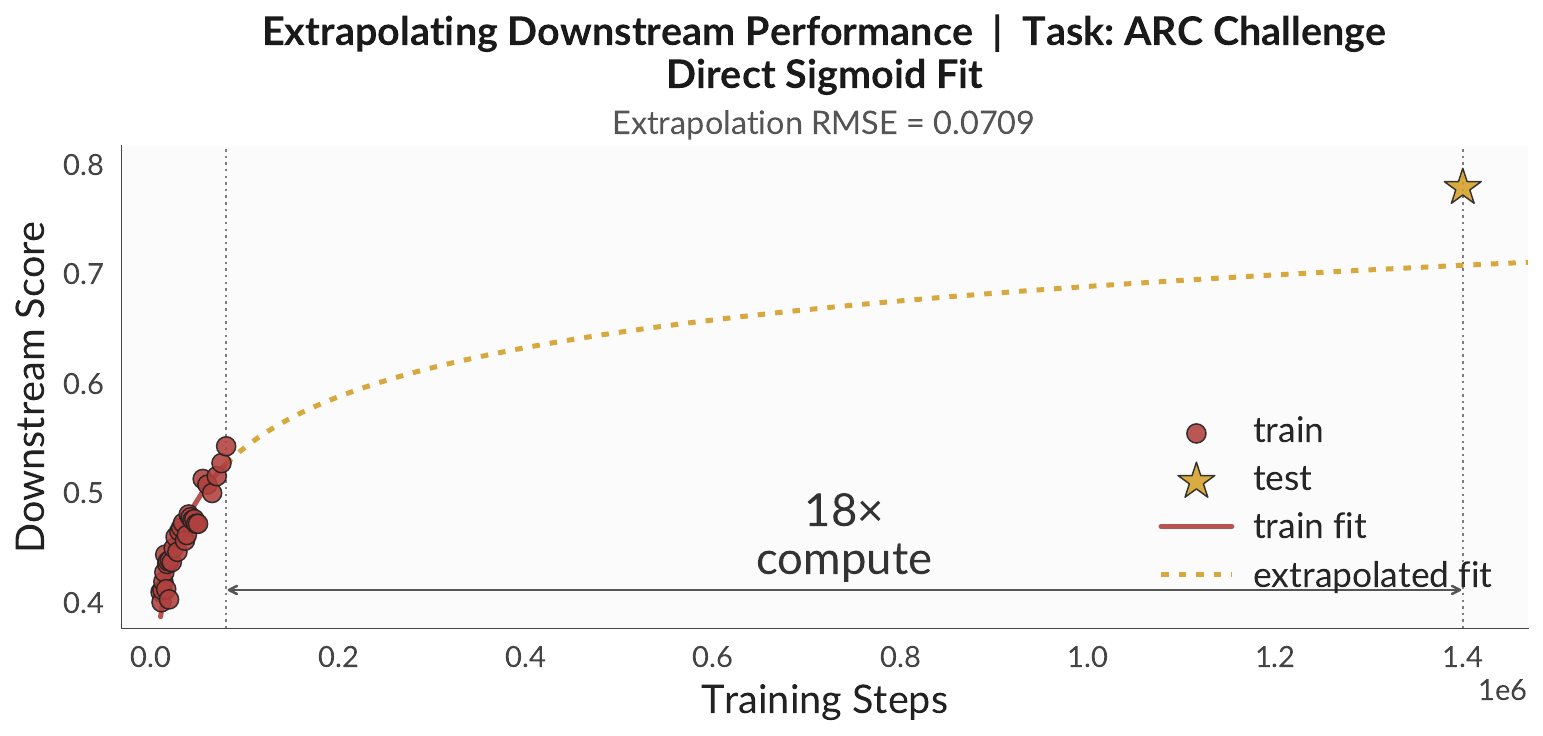}
\end{subfigure}
\hfill
\begin{subfigure}[t]{0.325\textwidth}
    \includegraphics[width=\textwidth]{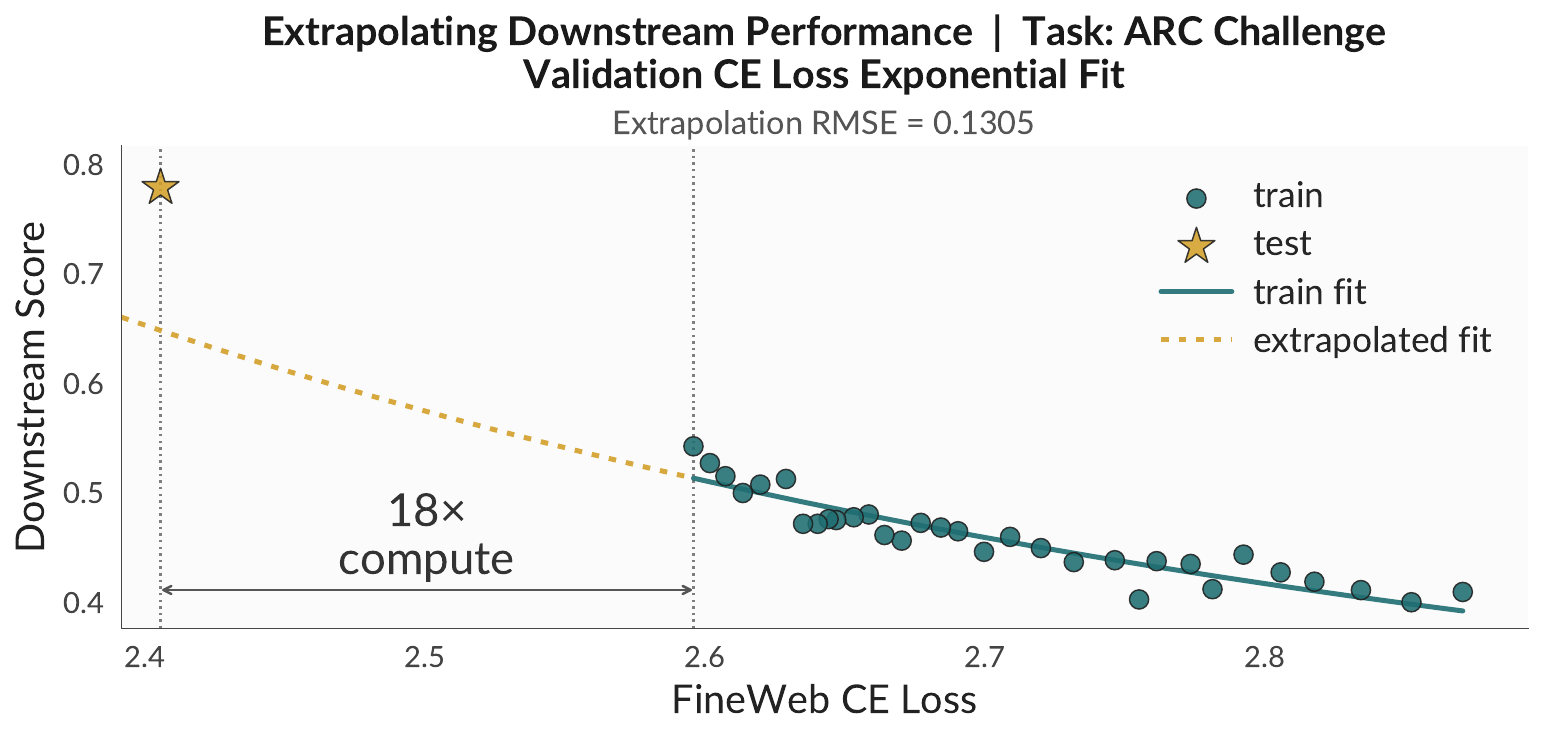}
\end{subfigure}
\hfill
\begin{subfigure}[t]{0.325\textwidth}
    \includegraphics[width=\textwidth]{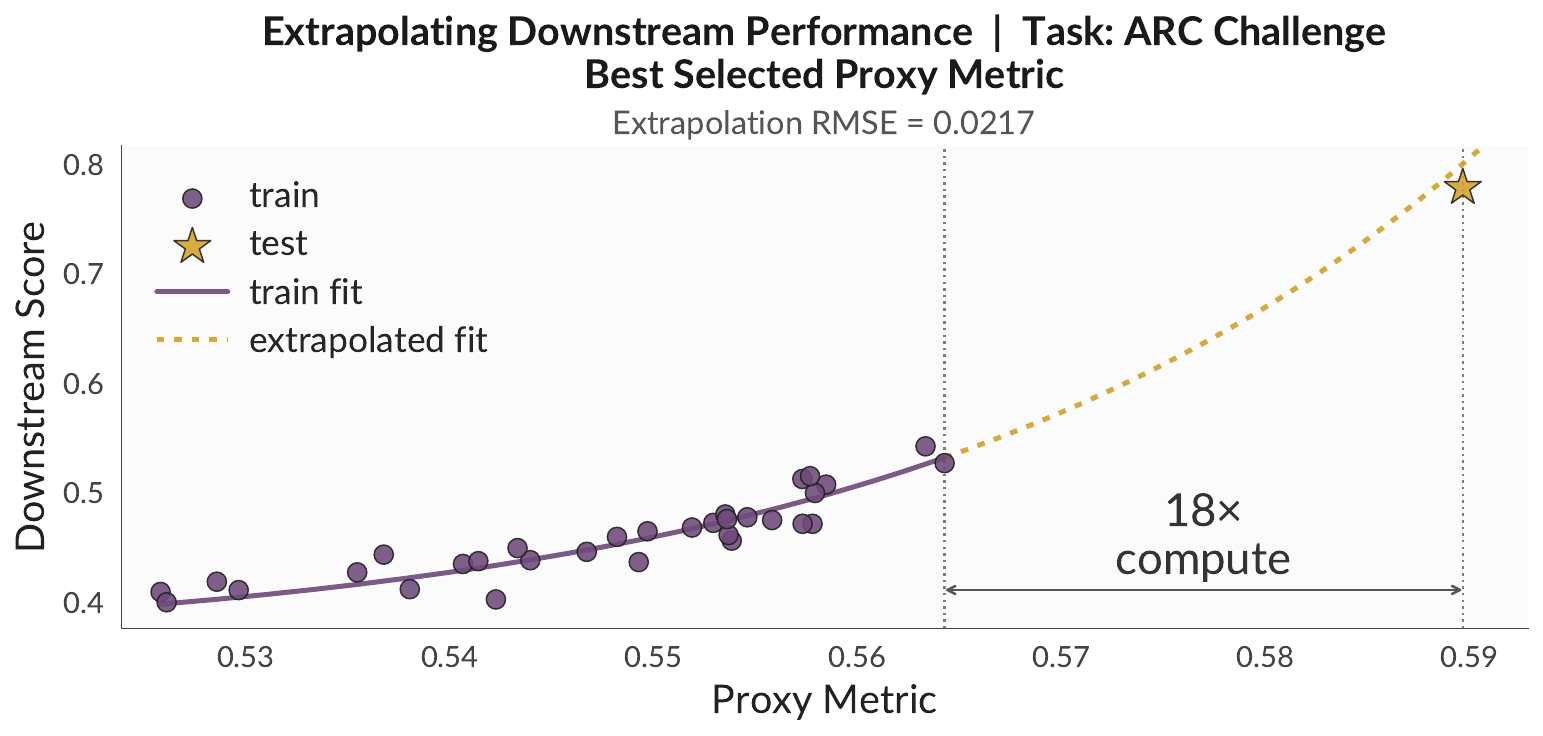}
\end{subfigure}
\caption{\textbf{Extrapolating ARC Challenge accuracy along the pretraining trajectory of OLMo-3-7B.} Circles are the training window (up to $80$K steps), the star is the held-out checkpoint at $1.4$M steps (${\sim}18\times$ the training compute). \emph{Left:} accuracy vs. $\log_{10}(\text{steps})$, fit with a sigmoid (RMSE $= 0.07$). \emph{Centre:} accuracy vs.\ CE loss on FineWeb, fit with an exponential (RMSE $= 0.13$). \emph{Right:} accuracy vs.\ the best univariate proxy, fit with a power law (RMSE $= 0.02$).}
\label{fig:proxy_vs_downstream_arc_challenge}
\end{figure}

\begin{table}[t]
\centering
\caption{Extrapolation RMSE for predicting downstream accuracy at $1.4$M steps from checkpoints up to $80$K steps (${\sim}18\times$ compute extrapolation). Three predictors are compared: the best univariate proxy (selected by inner split), an exponential fit of accuracy against FineWeb validation CE loss \citep{gadre2025language}, and a direct sigmoid fit of accuracy against $\log_{10}(\text{steps})$ \citep{owen2024predictablelanguagemodelbenchmark}. \textbf{Bold} marks the lowest RMSE per task.}
\label{tab:proxy_vs_downstream}
\setlength{\tabcolsep}{6pt}
\renewcommand{\arraystretch}{1.15}
\begin{tabular}{llccc}
\toprule
Task & Best Proxy & Proxy & CE Loss & Compute \\
\midrule
HellaSwag      & \texttt{uniform\,/\,top\_5\_acc}              & \textbf{0.003} & 0.094 & 0.105 \\
BoolQ          & \texttt{entropy\,/\,top\_5\_acc}              & 0.037 & 0.111 & \textbf{0.034} \\
SocialIQA      & \texttt{inverse\_freq\,/\,top\_3\_acc}   & 0.016 & \textbf{0.011} & 0.025 \\
Winogrande     & \texttt{inverse\_freq\,/\,top\_2\_acc}   & \textbf{0.010} & 0.078 & 0.018 \\
ARC Challenge  & \texttt{expert\_disagreement\,/\,top\_3\_acc}             & \textbf{0.022} & 0.131 & 0.071 \\
ARC Easy       & \texttt{probability\,/\,top\_3\_acc}          & 0.049 & \textbf{0.001} & 0.036 \\
CommonsenseQA  & \texttt{inverse\_freq\,/\,top\_1\_acc}   & \textbf{0.004} & 0.023 & 0.129 \\
MMLU           & \texttt{freq\,/\,top\_2\_acc}            & \textbf{0.001} & 0.036 & 0.058 \\
OpenBookQA     & \texttt{entropy\,/\,top\_1\_acc}              & 0.069 & 0.086 & \textbf{0.000} \\
PIQA           & \texttt{probability\,/\,reciprocal\_rank}         & 0.029 & \textbf{0.025} & 0.068 \\
\midrule
\textbf{Mean}  & ---                                               & \textbf{0.024} & 0.059 & 0.055 \\
\bottomrule
\end{tabular}
\end{table}

\subsection{Training-time Forecasting}
\label{app:extrapolation}

This section provides some additional results for the training-time forecasting experiments of \S\ref{sec:extrapolation}.

\paragraph{Pretraining extrapolation for AIME and SuperGPQA.}
\Cref{fig:extrapolation_pre_aime_supergpqa} extends the pretraining extrapolation of \cref{fig:extrapolation} (left) to the two benchmarks not shown in the main figure: AIME and SuperGPQA. On AIME the selected proxy follows a clean power law with NMAE $= 0.007$ and train-window $R^2 = 0.93$, and the extrapolated value lands almost exactly on the held-out checkpoint. SuperGPQA is similarly well-behaved (NMAE $= 0.035$, $R^2 = 0.97$), with a slightly larger extrapolation gap. Together with the four benchmarks in the main figure (mean NMAE $= 0.034$), these results confirm that the power-law regularity of the proxy is not confined to a subset of tasks: across all six reasoning benchmarks the mean pretraining NMAE is $0.030$.

\paragraph{Post-training extrapolation for AIME.}
\Cref{fig:extrapolation_post_aime} shows the post-training extrapolation on AIME, complementing the set of results in \cref{fig:extrapolation} (right). This is the noisiest setting we observe: the training-window points exhibit visible scatter around the fitted power law ($R^2 = 0.88$), and the NMAE is $0.110$, roughly three times larger than the post-training average on the other four benchmarks. We attribute this to the difficulty and noisiness of the task itself. AIME consists of only $30$ problems, all of which are competition-math level difficulty.

\paragraph{Proxy-accuracy correspondence at post-training checkpoints.}
The extrapolation experiments predict the \emph{proxy metric} at a future checkpoint. For this prediction to be useful, the proxy must actually track the downstream quantity of interest. \Cref{fig:extrapolation_post_scatter_usaco_hmmt} provides this sanity check by plotting the selected univariate proxy against downstream accuracy across post-training checkpoints of OLMo-3-7B-Think on USACO and HMMT. On USACO the relationship is strongly monotonic (Spearman $\rho = 0.91$). On HMMT the trend is noisier ($\rho = 0.83$) with wider confidence intervals in general, reflecting the smaller size of the test set and the high difficulty of competition mathematics. Both correlations are well above what would be needed for the extrapolated proxy to serve as a reliable early indicator of downstream ranking, and the mean across all five post-training benchmarks ($\rho = 0.84$, as reported in \S\ref{sec:extrap_proxy}) confirms that this correspondence holds across the other benchmarks.

\paragraph{Proxy metrics selected for per-task downstream accuracy extrapolation.}
\Cref{tab:proxy_vs_downstream} expands \cref{tab:proxy_vs_downstream_main} by reporting the identity of the best proxy selected by the inner-split procedure for each of the ten OLMES benchmarks. Two patterns stand out. First, the selected proxies are overwhelmingly top-$k$ accuracy variants: nine of the ten tasks select a top-$k$ accuracy metric under some weighting scheme, with $k$ ranging from $1$ to $5$. The one exception is PIQA, which selects probability-weighted reciprocal rank. This convergence on top-$k$ agreement is consistent with the cross-family heatmap in \cref{fig:feature_heatmap}, where top-$k$ accuracy under various weightings dominates the selected features, and suggests that how often a model places the expert's token in its top predictions is a broadly informative signal across diverse evaluation settings. Second, the weighting schemes are heterogeneous: inverse-frequency weighting appears on three tasks (Winogrande, CommonsenseQA, SocialIQA), frequency weighting on one (MMLU), entropy weighting on one (BoolQ), and other schemes on the remainder. No single weighting dominates, which is why the full library of eighty proxy metrics is necessary for broad coverage.

\paragraph{Baseline comparisons for per-task downstream accuracy extrapolation.}
\Cref{fig:sigmoid_hellaswag} shows the direct sigmoid baseline of \citet{owen2024predictablelanguagemodelbenchmark} on HellaSwag. Within the training window (up to $80$K steps) the sigmoid fits the data well, but at the held-out checkpoint ($1.4$M steps) it overshoots the true accuracy substantially (RMSE $= 0.11$). \Cref{fig:proxy_vs_downstream_winogrande,fig:proxy_vs_downstream_arc_challenge} provide side-by-side comparisons of all three predictors on Winogrande and ARC Challenge. On Winogrande the sigmoid fit undershoots slightly (RMSE $= 0.02$), the CE loss exponential overshoots markedly (RMSE $= 0.08$), and the proxy power law tracks the target closely (RMSE $= 0.01$). ARC Challenge exhibits the same pattern more dramatically: the CE loss exponential diverges (RMSE $= 0.13$), the sigmoid is closer but still off (RMSE $= 0.07$), while the proxy fit remains tight (RMSE $= 0.02$).

\section{Extended Related Work}
\label{app:extended_related}

In this section, we mention other related works.

\paragraph{Compute-based scaling laws for pretraining loss.}
Classical scaling laws connect training compute, parameters, and tokens to a single pretraining cross-entropy loss under controlled architectures \citep{kaplan2020scalinglawsneurallanguage,hoffmann2022an}. These laws are extraordinarily powerful as planning tools when the quantity of interest is loss itself, and the functional form, a power law with additive terms in $N$ and $D$, is the starting point for virtually every downstream forecasting pipeline that follows, including ours when we extrapolate the proxy metric along a training trajectory in \cref{sec:extrapolation}. They are silent, however, on the quantity practitioners actually care about, which is the score of the trained model on a specific downstream task. Follow-up work extends these laws to over-trained regimes \citep{gadre2025language}, data-constrained regimes \citep{muennighoff2023scaling}, inference-aware compute budgets \citep{sardana2024beyond}, and has sharpened their fitting protocols \citep{choshen2025hitchhiker, besiroglu2024chinchilla, porian2024resolving}. The proxy metrics we study complement this line by producing a dense, task-specific, and well-behaved scalar that is defined at every training step and obeys its own smooth law, even in regimes where pretraining loss has long since saturated its informativeness about task capability.

\paragraph{Forecasting downstream task performance.}
A growing line of work converts the pretraining loss into a forecast of downstream accuracy. \citet{gadre2025language} propose the exponential map $y = \epsilon - k \cdot \exp(-\gamma x)$ from validation perplexity to average downstream top-1 error across many tasks. \citet{bhagia2025establishing} decompose the problem into two stages, $(N, D) \to$ task-specific loss on the correct answer $\to$ ranked-classification accuracy, using compute-efficient ladders at roughly $1\%$ of target compute. \citet{chen2025scaling} extend the two-stage approach to domain-mixed pretraining with FLP and FLP-M. \citet{ruan2024observational} take a different route with observational scaling laws, fitting a low-dimensional latent capability axis over roughly one hundred public models via PCA on benchmark scores, and showing that agentic and reasoning behaviors become predictable in this latent space. \citet{owen2024predictablelanguagemodelbenchmark} fits sigmoidal forms from scaling-estimated loss to BIG-Bench and MMLU across eleven families, and \citet{polo2026sloth} show that benchmark performance is well explained by a handful of latent skills that transfer across families. \citet{krajewski2026revisiting} push the single-recipe direct law further by showing that a two-parameter law on training compute can fit downstream accuracy directly when the task is above chance.

\paragraph{The unreliability of downstream scaling laws under realistic constraints.}
Several recent papers document concrete failure modes of the scaling-law-to-downstream pipeline. \citet{lourie2025unreliable} find that only about $39\%$ of the tasks studied by \citet{gadre2025language} exhibit predictably linear scaling of downstream accuracy with compute, and \citet{schaeffer2025why} argue that predicting downstream capabilities of frontier models from scale has remained elusive because argmax-based accuracy degrades the cross-entropy-to-score relationship. Accurate prediction requires modeling probability mass on both correct and incorrect completions. This observation is a direct motivation for the margin and wrong-confidence features in our library, which measure exactly the two sides of this error surface. \citet{isik2025scaling} provide a complementary warning. On machine translation, pretraining cross-entropy can continue to fall while BLEU or COMET degrade, so a scalar loss is not even a monotone predictor of the downstream metric. \citet{liu2023same} show that models with identical pretraining loss can differ substantially downstream, a direct counterexample to the simplest loss-as-rank hypothesis, and \citet{tay2023architectures} show that the upstream-downstream ordering can invert across architectures.

\paragraph{Emergence and the role of continuous metrics.}
\citet{wei2022emergent} coined the term emergent abilities for capabilities that appear absent in smaller models and abruptly present at larger scales, a framing that was interpreted by many as evidence against forecastability. \citet{schaeffer2023are} push back by showing that apparent emergence largely evaporates under continuous and smooth metrics such as log-likelihood, Brier score, or edit distance, with abrupt jumps appearing only when the evaluation collapses the predictive distribution to a discrete score. \citet{du2024loss_emergence} strengthen this picture from the opposite direction by arguing that downstream abilities align more cleanly with pretraining loss than with parameter count, and that emergence often reflects a loss threshold. \citet{hu2024predicting} develop PassUntil, a continuous evaluation with effectively infinite resolution that makes per-instance task-solve probability predictable from small-scale models, and \citet{snell2024predicting} fit emergence laws by finetuning smaller models to shift the emergence point.

\paragraph{Loss-to-loss and cross-distribution prediction.}
\citet{brandfonbrener2024l2l} introduce loss-to-loss prediction, fitting shifted power laws that map training loss on one dataset to training or test loss on another and extrapolating up to roughly twenty times the training compute. \citet{mayilvahanan2025llms} extend this to thousands of checkpoints, showing that the loss-to-loss trend is robust across architectures and tokenizers. \citet{huang2024compression} provide cross-family evidence in the same spirit, reporting a Pearson correlation of approximately $-0.95$ between bits-per-character and downstream scores across thirty-one LLMs and twelve benchmarks. \citet{magnusson2024paloma} propose Paloma as a tokenizer-invariant bits-per-byte benchmark over 546 domains, and \citet{xia2023training} show that across the OPT family perplexity predicts in-context learning on seventy-four BIG-Bench tasks better than model size or compute.

\paragraph{Token-level likelihood signals and fine-grained weighting.}
Several papers have recognized that not every token position carries the same signal. \citet{gonen2023demystifying} show that prompt perplexity is a label-free predictor of task performance, \citet{ankner2025perplexed} use small-reference-model perplexity to prune pretraining data and improve downstream accuracy at 3B scale, and \citet{thrush2025improving} select pretraining documents whose per-document perplexity correlates most strongly with downstream benchmarks. Closest in spirit to our weighting schemes, \citet{fang2025what} introduce LongPPL, which weights perplexity only on key tokens identified by a reference model and achieves a $-0.96$ correlation with long-context benchmarks where uniform perplexity shows almost none. \citet{sorscher2022neural} use the teacher's margin (top minus second-top logit) as a per-example difficulty signal for data pruning, anticipating our use of margin as a token-level feature.

\paragraph{Small-scale proxies for pretraining decisions.}
Another line of works asks whether small models can be used to decide between candidate pretraining datasets or recipes before committing target-scale compute. \citet{wortsman2024smallscale} formalize this as small-scale proxies for large-scale transformer behavior, and $\mu$P and $\mu$Transfer \citep{yang2021tuning} provide the parameterization that makes such extrapolation work for hyperparameters. \citet{xie2023doremi} introduce DoReMi, which trains a 280M proxy with Group DRO on excess NLL and transfers domain weights to 8B. \citet{liu2025regmix} introduce RegMix, which fits a regression over 512 1M-parameter proxies to select optimal mixtures at 1B and 7B with roughly $2\%$ of full compute. \citet{magnusson2025datadecide} construct DataDecide, a large controlled testbed of twenty-five pretraining corpora and fourteen proxy scales, and show that continuous likelihood-style metrics make MMLU, ARC, HellaSwag, MBPP, and HumanEval over $80\%$ predictable at $0.01\%$ of target compute. \citet{koh2025rbridge} propose rBridge, which aligns the small-proxy NLL with the target task by weighting by a frontier-model reasoning trace.

\paragraph{Reasoning traces as supervision and as signal.}
Reasoning traces have been used most extensively as a training signal. Chain-of-thought prompting \citep{wei2022cot} and self-consistency \citep{wang2022selfconsistency} exploit reasoning trajectories at inference, and STaR \citep{zelikman2022star} and Quiet-STaR \citep{zelikman2024quietstar} bootstrap models from generated rationales. Process reward models make the token-level view explicit. \citep{uesato2022solvingmathwordproblems} compare process and outcome supervision, \citet{lightman2024lets} release PRM800K and show that process supervision dominates outcome supervision, and \citet{wang2024mathshepherd} and \citep{luo2024improvemathematicalreasoninglanguage} develop automatic step-level process rewards from Monte Carlo rollouts. Two recent methods are particularly close in spirit to ours at the scoring step. \citet{hao2023reasoning} use per-step LM token likelihood together with self-evaluation as an MCTS reward, and \citet{cui2025prime} and \citet{yuan2024free} derive implicit token-level process rewards from log-likelihood ratios of outcome-supervised models. All of these works treat trajectories as signals for training or search. Our paper uses trajectories as an evaluation substrate instead.



\end{document}